\documentclass[10pt, logo, twocolumn, copyright]{nvidiatechreport}

\usepackage{natbib}
\usepackage{hyperref}
\usepackage{url}
\usepackage{booktabs}
\usepackage{amsmath}
\usepackage{amssymb}
\usepackage{mathtools}
\usepackage{amsthm}
\usepackage{enumitem}
\usepackage{multirow}
\usepackage{capt-of}
\usepackage{wrapfig}
\usepackage{bm}
\usepackage{bbm}
\usepackage{arydshln}
\usepackage{placeins}
\usepackage{comment}
\usepackage{algorithm}
\usepackage{algorithmic}
\usepackage{eqparbox}
\usepackage{colortbl}
\usepackage{tikz}
\usepackage{makecell}
\usepackage{adjustbox}
\usepackage{tabularx}
\usepackage{xspace}
\usepackage{subcaption}

\title{GenMol: A Drug Discovery Generalist with Discrete Diffusion}

\author{
Seul Lee\texttwosuperior$^*$,
Karsten Kreis\textonesuperior,
Srimukh Prasad Veccham\textonesuperior,
Meng Liu\textonesuperior,
Danny Reidenbach\textonesuperior,
Yuxing Peng\textonesuperior,
Saee Paliwal\textonesuperior,
Weili Nie\textonesuperior$^\dagger$,
Arash Vahdat\textonesuperior$^\dagger$ \\
\textonesuperior\ NVIDIA \, \texttwosuperior\ KAIST}

\correspondingauthor{X}

\begin{abstract}

\textbf{Abstract:} % for NVIDIA technical report
Drug discovery is a complex process that involves multiple stages and tasks. However, existing molecular generative models can only tackle some of these tasks. We present \emph{Generalist Molecular generative model} (GenMol), a versatile framework that uses only a \emph{single} discrete diffusion model to handle diverse drug discovery scenarios. GenMol generates Sequential Attachment-based Fragment Embedding (SAFE) sequences through non-autoregressive bidirectional parallel decoding, thereby allowing the utilization of a molecular context that does not rely on the specific token ordering while having better sampling efficiency. GenMol uses fragments as basic building blocks for molecules and introduces \emph{fragment remasking}, a strategy that optimizes molecules by regenerating masked fragments, enabling effective exploration of chemical space. We further propose \emph{molecular context guidance} (MCG), a guidance method tailored for masked discrete diffusion of GenMol. GenMol significantly outperforms the previous GPT-based model in \emph{de novo} generation and fragment-constrained generation, and achieves state-of-the-art performance in goal-directed hit generation and lead optimization. These results demonstrate that GenMol can tackle a wide range of drug discovery tasks, providing a unified and versatile approach for molecular design. Our code is available at \url{https://github.com/NVIDIA-Digital-Bio/genmol}.
\end{abstract}

\begin{document}
\maketitle

\section{Introduction}

Discovering molecules with the desired chemical profile is the core objective of drug discovery~\citep{hughes2011principles}. To achieve the ultimate goal of overcoming disease, a variety of drug discovery approaches have been established. For example, fragment-constrained molecule generation is a popular strategy for designing new drug candidates under the constraint of preserving a certain molecular substructure already known to exhibit a particular bioactivity~\citep{murray2009rise}. Furthermore, real-world drug discovery pipelines are not a single stage but consist of several key stages, such as hit generation and lead optimization~\citep{hughes2011principles}. A drug discovery process that leads to the finding of drug candidates that can enter clinical trials should consider all of these different scenarios.

\begin{figure}[t!]
    \centering
    \includegraphics[width=0.6\linewidth]{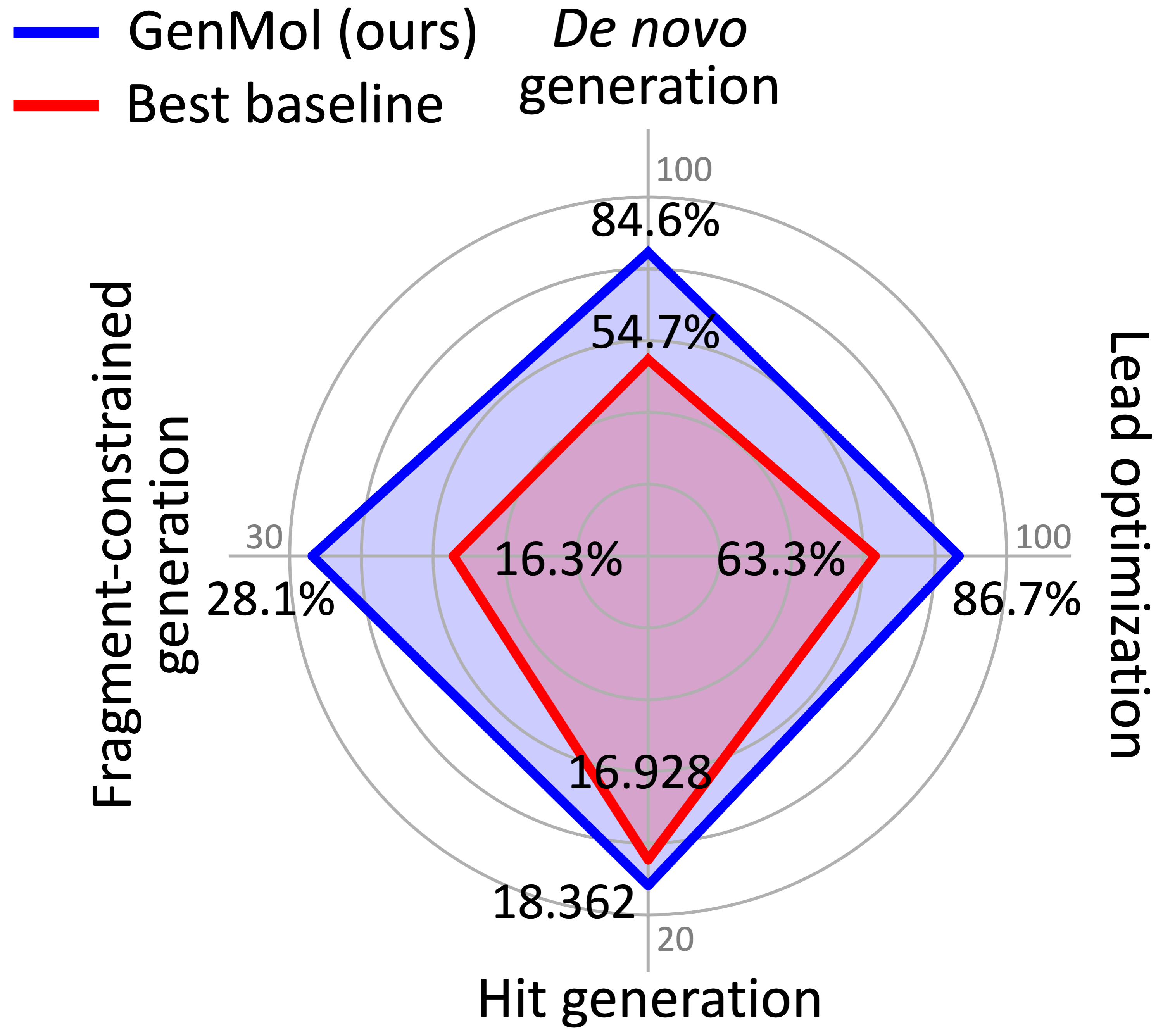}
    \vspace{-0.05in}
    \caption{\small \textbf{Results on drug discovery tasks.} The values are quality, average quality, sum AUC top-10, and success rate for \emph{de novo} generation, fragment-constrained generation, hit generation, and lead optimization, respectively. The ``best baseline'' refers to multiple best-performing task-specific models among prior works.}
    \label{fig:radar}
    \vspace{-0.1in}
\end{figure}

\begin{figure*}[t!]
    \centering
    \includegraphics[width=\linewidth]{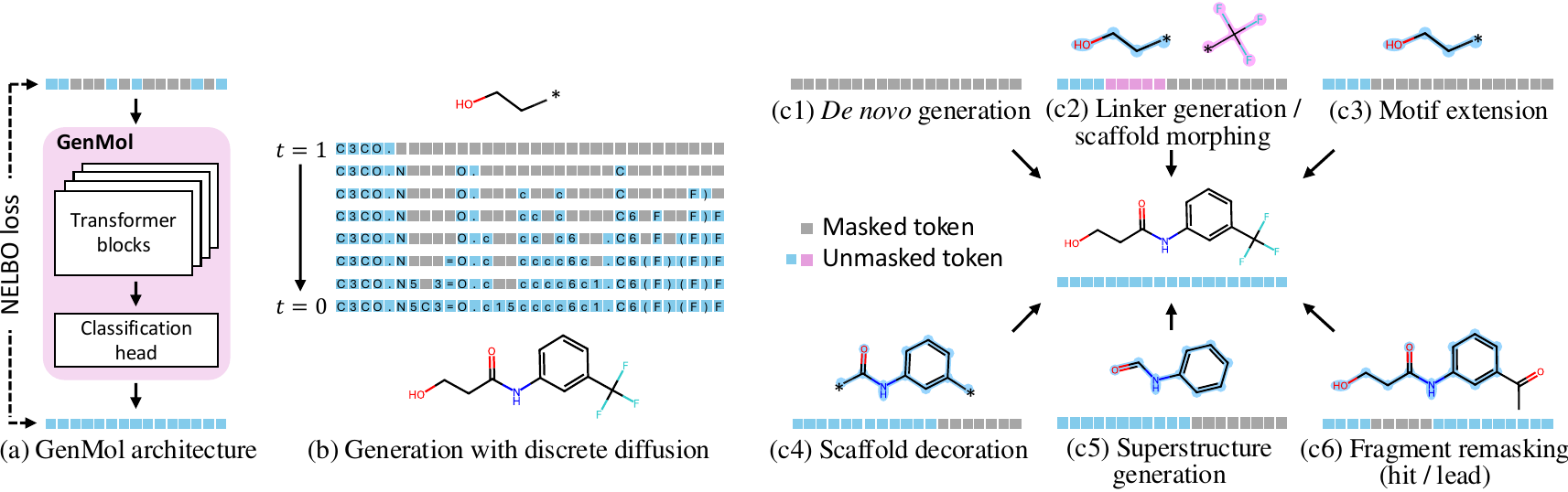}
    \vspace{-0.2in}
    \caption{\small \textbf{(a) GenMol architecture.} GenMol adopts the BERT architecture and is trained with the NELBO loss of masked discrete diffusion. \textbf{(b) Generation process of GenMol.} Under masked discrete diffusion, GenMol completes a molecule by simulating backward in time and predicting masked tokens at each time step $t$ until all tokens are unmasked. \textbf{(c) Illustration of various drug discovery tasks that can be performed by GenMol.} GenMol is endowed with the ability to easily perform (c1) \emph{de novo} generation, (c2-c5) fragment-constrained generation, and (c6) fragment remasking that can be applied to goal-directed hit generation and lead optimization.}
    \label{fig:concept}
    \vspace{-0.1in}
\end{figure*}

Generative models have emerged as a promising methodology to accelerate labor-intensive drug discovery pipelines~\citep{olivecrona2017molecular,jin2018junction,yang2021hit,lee2023exploring}, but previous molecular generative models have a common limitation: they focus on only one or two of the drug discovery scenarios. They either cannot be applied to multiple tasks or require expensive modifications including retraining of a specific architecture for each task~\citep{yang2020syntalinker,guo2023link}. Recently, SAFE-GPT~\citep{noutahi2024gotta} has been proposed to address this problem by formulating several molecular tasks as a fragment-constrained generation task, solved by sequence completion. SAFE-GPT uses Sequential Attachment-based Fragment Embedding (SAFE) of molecules, which represents a molecule as an unordered sequence of Simplified Molecular Input Line Entry System (SMILES)~\citep{weininger1988smiles} fragment blocks. However, since all sequence-based molecular representations including SMILES and SAFE assume an ordering of tokens based on heuristic rules such as depth-first search (DFS), autoregressive models that operate in a left-to-right order like GPT are unnatural for processing and generating molecular sequences.
In addition, the autoregressive decoding scheme limits the computational efficiency of the model and makes it challenging to introduce guidance during generation. Moreover, SAFE-GPT relies on finetuning under expensive reinforcement learning (RL) objectives to be applied to goal-directed molecule generation.
\looseness=-1

To tackle these limitations, we propose \emph{Generalist Molecular generative model} (GenMol), a versatile molecular generation framework that is endowed with the ability to handle diverse scenarios that can be encountered in the multifaceted drug discovery pipeline (Figure~\ref{fig:concept}(c)). GenMol adopts a masked discrete diffusion framework~\citep{austin2021structured,sahoo2024simple,shi2024simplified} with the BERT architecture~\citep{devlin2018bert} to generate SAFE molecular sequences, thereby enjoying several advantages: (i) Discrete diffusion allows GenMol to exploit a molecular context that does not rely on the specific ordering of tokens and fragments by bidirectional attention. (ii) The non-autoregressive parallel decoding improves GenMol's computational efficiency (Figure~\ref{fig:concept}(b)). (iii) Discrete diffusion enables GenMol to explore chemical space with a simple yet effective remasking strategy. We propose \emph{fragment remasking} (Figure~\ref{fig:concept}(c6) and Figure~\ref{fig:remasking}), a strategy to optimize molecules by replacing certain fragments in a given molecule with masked tokens, from which diffusion generates new fragments. Utilizing fragments as the explorative unit instead of individual tokens is more in line with chemists' intuition during optimization in drug discovery, and GenMol can effectively and efficiently explore the vast chemical space to find chemical optima. (iv) Discrete diffusion also makes it possible to apply guidance during generation based on the entire sequence. To this end, we propose \emph{molecular context guidance} (MCG), a guidance method to improve the performance of GenMol by calibrating its predictions with information in a given molecular context.

We experimentally validate GenMol on a wide range of molecule generation tasks that simulate real-world drug discovery problems, including \emph{de novo} generation, fragment-constrained generation, goal-directed hit generation, and goal-directed lead optimization. Across extensive experiments, GenMol outperforms existing methods by a large margin (Figure~\ref{fig:radar}).
Note that the best baseline results shown in Figure~\ref{fig:radar} are not the results of a single model, but of multiple task-specific models. These results demonstrate GenMol's potential as a versatile tool that can be used throughout the drug discovery pipeline.
\looseness=-1

We summarize our contributions as follows:
\begin{itemize}[itemsep=0mm, parsep=2pt, leftmargin=15pt]
    \vspace{-0.12in}
    \item We introduce GenMol, a framework for unified and versatile molecule generation by building masked discrete diffusion that generates SAFE molecular sequences.
    \item We propose fragment remasking, an effective strategy for exploring chemical space using molecular fragments as the unit of exploration.
    \item We propose MCG, a guidance scheme for GenMol to effectively utilize molecular context information.
    \item We validate the efficacy and versatility of GenMol on a wide range of drug discovery tasks.
\end{itemize}

\section{Related Work}

\paragraph{Discrete diffusion.}
There has been steady progress in applying discrete diffusion for discrete data generation, especially in NLP tasks~\citep{hoogeboom2021argmax,austin2021structured,he2022diffusionbert,zheng2023reparameterized,lou2023discrete,sahoo2024simple}. This is mainly due to their non-autoregressive generation property, which leads to a potential for better modeling long-rang bidirectional dependencies and acclerating sampling speed, and their flexible design choices in training, sampling, and controllable generation~\citep{sahoo2024simple}. 
Notably, D3PM~\citep{austin2021structured} introduced a general framework with a Markov forward process represented by transition matrices, and a transition matrix with an absorbing state corresponds to the masked language modeling (MLM) such as BERT~\citep{devlin2018bert}. \citet{campbell2022continuous} proposed a continuous-time framework for discrete diffusion models based on the continuous-time Markov chain (CTMC) theory. SEDD~\citep{lou2023discrete} introduced a denoising score entropy loss that extends score matching to discrete diffusion models. 
\citet{sahoo2024simple} and \citet{shi2024simplified} proposed simple masked discrete diffusion frameworks, with the training objective being a weighted average of MLM losses across different diffusion time steps.
\looseness=-1

Recently, a few works have applied discrete diffusion for molecular generation. For instance, DiGress~\citep{vignac2022digress} followed the D3PM framework to generate molecular graphs with categorical node and edge attributes. Other works~\citep{zhang2023diffmol,lin2024diffbp,hua2024mudiff} focused on the 3D molecular structure generation, where they used discrete diffusion for atom type generation and  continuous diffusion for atom position generation. However, none of them applied discrete diffusion for molecular sequence generation that can serve as a generalist foundation model for solving various downstream tasks.

\vspace{-0.05in}
\paragraph{Fragment-based drug discovery.}
Fragment-based molecular generative models refer to a class of methods that reassemble existing molecular substructures (i.e., fragments) to generate new molecules.
They have been consider as an effective drug discovery approach as (i) assembling fragments simplifies the generation process and improves chemical validity and (ii) the unit that determines biochemical effect of a molecule is a fragment rather than an individual atom~\citep{li2020application}.
A line of works~\citep{jin2020multi,maziarz2021learning,kong2022molecule,geng2023novo} used graph-based VAEs to generate novel molecules conditioned on discovered substructures.  
\citet{xie2020mars} proposed to progressively add or delete fragments of molecular graphs using Markov chain Monte Carlo (MCMC) sampling. 
\citet{yang2021hit} and \citet{powers2023geometric} used a reinforcement learning (RL) framework and classification, respectively, to progressively add fragments to the incomplete molecule.
Graph-based genetic algorithms (GAs)~\citep{jensen2019graph,tripp2023genetic} is a strong approach that decomposes parent molecules into fragments that are combined to generate an offspring molecule. 
However, since their generation is from random combinations of existing fragments with a local mutation of a small probability, they suffer from limited exploration in the chemical space.
More recently, $f$-RAG~\citep{lee2024frag} introduced a fragment-level retrieval framework that augments the pre-trained molecular language model SAFE-GPT~\citep{noutahi2024gotta}, where retrieving fragments from dynamically updated fragment vocabulary largely improves the exploration-exploitation trade-off.
However, $f$-RAG still needs to train an information fusion module before adapting to various goal-oriented generation tasks.
\looseness=-1

\section{Background}

\subsection{Masked Diffusion~\label{sec:background_mdlm}}

Masked diffusion models~\citep{sahoo2024simple,shi2024simplified} are a simple and effective class of discrete diffusion models, and we follow MDLM~\citep{sahoo2024simple} to define our masked diffusion. Formally, we define $\pmb{x}$ as a sequence of $L$ tokens, each of which, denoted as $\pmb{x}^{l}$, is a one-hot vector with $K$ categories (i.e., $\pmb{x}^{l}_i \in \{0, 1\}^K$ and $\sum_{i=1}^K \pmb{x}^{l}_i = 1$). Without loss of generality, we assume the $K$-th category represents the masking token, whose one-hot vector is denoted by $\textbf{m}$ (i.e., $\textbf{m}_K = 1$). We also define $\text{Cat}(\cdot;\pmb{\pi})$ as a categorical distribution with a probability $\pmb{\pi} \in \Delta^K$, where $\Delta^K$ represents the simplex over $K$ categories.
\looseness=-1

The forward masking process independently interpolates the probability mass between each token in clean data sequence $\pmb{x}^{l}$ and the masking token $\textbf{m}$, defined as
\begin{align}
    q(\pmb{z}^{l}_t | \pmb{x}^{l}) = \text{Cat}(\pmb{z}^{l}_t; \alpha_t \pmb{x}^{l} + (1-\alpha_t) \textbf{m}),
    \label{eq:fwd_masked}
\end{align}
where $\pmb{z}^{l}_t$ denotes the $l$-th token in the noisy data sequence at the time step $t \in [0, 1]$, and $\alpha_t \in [0,1]$ denotes the masking ratio that is monotonically decreasing function of $t$, with $\alpha_0=1$ to $\alpha_1=0$. Accordingly, at time step $t=1$, $\pmb{z}_t$ becomes a sequence of all masked tokens.

The reverse unmasking process inverts the masking process and independently infers each token of the less masked data $\pmb{z}_s$ from more masked data $\pmb{z}_t$ with $s<t$, which is given by
\begin{align}
    \fontsize{9pt}{9pt}
    \hspace{-5mm}
    p_\theta(\pmb{z}^{l}_s | \pmb{z}^{l}_t) = 
    \begin{cases}
        \text{Cat}(\pmb{z}^{l}_s; \pmb{z}^{l}_t) &\pmb{z}^{l}_t \neq \textbf{m} \\
        \text{Cat}(\pmb{z}^{l}_s; \frac{ (1-\alpha_s)\textbf{m} + (\alpha_s - \alpha_t) \pmb{x}^{l}_\theta(\pmb{z}_t, t) }{1-\alpha_t}) &\pmb{z}^{l}_t = \textbf{m},
    \end{cases}
    \hspace{-5mm}
    \label{eq:rev_masked}
\end{align}
where $\pmb{x}_\theta(\pmb{z}_t, t)$ is a denoising network that takes the noisy data sequence $\pmb{z}_t$ as input and predicts $L$ probability vectors for the clean data sequence. 
This parameterization designs the reverse process such that it does not change unmasked tokens. 
To train the denoising network $\pmb{x}_\theta(\pmb{z}_t, t)$, the training objective, which implicitly approximates the negative ELBO~\citep{sohl2015deep}, is given by
\begin{align}
    \mathcal{L}_{\text{NELBO}} = \mathbb{E}_q \int_{0}^1 \frac{\alpha'_t}{1-\alpha_t} \sum_l \log \langle \pmb{x}^{l}_\theta(\pmb{z}_t, t), \pmb{x}^{l} \rangle \text{d}t,
    \label{eq:obj_masked}
\end{align}
which is the weighted average of MLM losses (i.e., cross-entropy losses) over time steps.

\subsection{SAFE Molecular Representation}

Simplified Molecular Input Line Entry System (SMILES)~\citep{weininger1988smiles} is the most widely used molecular string representation, but it relies on a heuristic depth-first search (DFS) that traverses the atoms of a molecule. Therefore, atoms that are close in molecular structure can be tokens that are very far apart in molecular sequence, and thus it is not straightforward to perform fragment-constrained molecular generation with SMILES.

Sequential Attachment-based Fragment Embedding (SAFE)~\citep{noutahi2024gotta} has been proposed to alleviate this problem. SAFE represents molecules as an unordered sequence of fragment blocks, thereby casting molecular design into sequence completion. SAFE is a non-canonical SMILES in which the arrangement of SMILES tokens corresponding to the same molecular fragment is consecutive. Molecules are decomposed into fragments by the BRICS algorithm~\citep{brics} and the fragments are concatenated using a dot token (``.'') while preserving their attachment points. SAFE is permutation-invariant on fragments, i.e., the order of fragments within a SAFE string does not change the molecular identity.
\looseness=-1

\section{Method}

\begin{figure*}[t!]
    \centering
    \includegraphics[width=\linewidth]{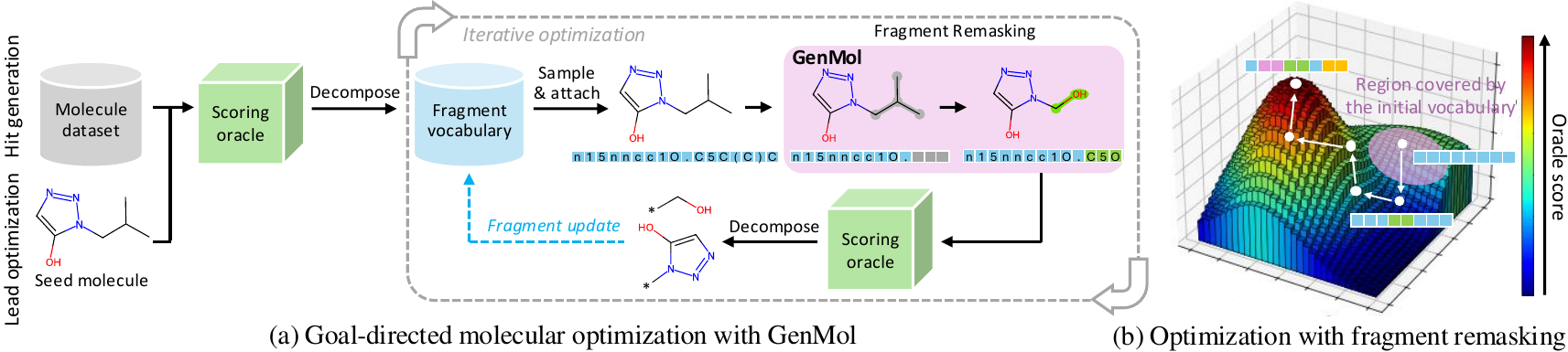}
    \vspace{-0.2in}
    \caption{\small \textbf{(a) Goal-directed hit generation and lead optimization process with GenMol.} An initial fragment vocabulary is constructed by decomposing an existing molecular dataset (hit generation) or a seed molecule (lead optimization). Two fragments are randomly sampled from the vocabulary and attached, and GenMol performs fragment remasking. The fragment vocabulary is updated with the generated molecules for the next iteration. \textbf{(b) Illustration of the molecular optimization trajectory with fragment remasking.} With fragment remasking, GenMol can explore beyond the initial fragment vocabulary to find chemical optima.}
    \label{fig:remasking}
    \vspace{-0.1in}
\end{figure*}

We introduce GenMol, a universal molecule generation framework that can solve various drug discovery tasks. We first introduce the construction of the discrete diffusion framework on the SAFE representation in Section~\ref{sec:method_diffusion}. Next, we describe the goal-oriented exploration strategy of GenMol with fragment remasking in Section~\ref{sec:method_remasking}. Finally, we describe MCG, a guidance scheme of GenMol by partially masking the given molecular context, in Section~\ref{sec:method_guidance}.

\subsection{Masked Diffusion for Molecule Generation~\label{sec:method_diffusion}}

We adopt masked discrete diffusion to generate SAFE sequences and establish a flexible and efficient molecule generation framework. Concretely, GenMol uses the BERT architecture~\citep{devlin2018bert} as the denoising network $\pmb{x}_\theta$ and the training scheme of MDLM. Utilizing discrete diffusion instead of an autoregressive model is more in line with the SAFE representation and has several advantages. First, due to the bidirectional attention in BERT, GenMol can leverage parallel decoding where all tokens are decoded simultaneously under discrete diffusion (Figure~\ref{fig:autoregressive}). As SAFE is fragment order-insensitive, this allows GenMol to predict masked tokens without relying on a specific ordering of generation while considering the entire molecule. The non-autoregressive parallel decoding scheme also improves GenMol's efficiency. Furthermore, the discrete diffusion framework enables GenMol to explore the neighborhood of a given molecule with a remasking strategy.

At each masked index $l$, GenMol samples $\pmb{z}^l_s$ based on the reverse unmasking process $p^l_\theta := p_\theta(\pmb{z}^l_s | \pmb{z}_t)$ specified by:
\begin{align}
    % \fontsize{9pt}{9pt}
    \fontsize{7pt}{7pt} % for NVIDIA report
    \hspace{-3mm}
    p^l_{\theta,i} = \frac{\exp\left(\log\pmb{x}^l_{\theta,i}(\pmb{z}_t,t)/\tau\right)}{\sum^K_{j=1}\exp\left(\log\pmb{x}^l_{\theta,j}(\pmb{z}_t,t)/\tau\right)} \text{ for } i=1,\cdots,K,
    \hspace{-2mm}
    \label{eq:p_l_i}
\end{align}
where $\log\pmb{x}^l_{\theta,i}(\pmb{z}_t,t)$ is the logit predicted by the model and $\tau$ is the softmax temperature. All masked tokens are predicted in a parallel manner and GenMol confirms the top-$N$ confident predictions with additional randomness $r$ following \citet{chang2022maskgit}, where $N$ is the number of tokens to unmask at each time step. Trade-offs between molecular quality and diversity often arise in drug discovery, and GenMol can balance them through the softmax temperature $\tau$ and the randomness $r$. Further details about confidence-based sampling is provided in Section~\ref{sec:confidence_sampling}.

\subsection{Exploring Chemical Space with GenMol~\label{sec:method_remasking}}

To perform goal-directed molecular optimization tasks, we propose a simple yet effective generation method (Figure~\ref{fig:remasking}) that consists of three steps: (1) fragment scoring, (2) fragment attaching, and (3) fragment remasking.

\paragraph{Fragment scoring.}
We start with constructing a fragment vocabulary. A set of $D$ molecules $\mathcal{D}=\{\big(\pmb{x}_d,y(\pmb{x}_d)\big)\}^D_{d=1}$, where $y(\pmb{x}_d)$ is the target property of molecule $\pmb{x}_j$, is decomposed into a set of $F$ fragments $\mathcal{F}=\{\pmb{f}_k\}_{k=1}^F$ using a predefined decomposition rule. We define the score of fragment $\pmb{f}_k$ following \citet{lee2024frag} as:
\looseness=-1
\begin{equation}
    y(\pmb{f}_k) = \frac{1}{|\mathcal{S}(\pmb{f}_k)|}\sum_{\pmb{x} \in \mathcal{S}(\pmb{f}_k)}y(\pmb{x}),
    \label{eq:score}
\end{equation}
where $\mathcal{S}(\pmb{f}_k)=\{\pmb{x}:\pmb{f}_k \text{ is a subgraph of } \pmb{x}\}$ and the top-$V$ fragments based on Eq.~(\ref{eq:score}) are selected as the vocabulary.

\paragraph{Fragment attaching.}
During generation, a molecule $\pmb{x}_\text{init}$ is first generated by randomly selecting two fragments from the vocabulary and attaching them. Fragments or functional groups influence the chemical properties of a molecule and therefore, fragments that commonly occur in molecules with desirable properties are likely to carry them to new molecules.
However, with fragment attaching alone, the model cannot generate new fragments that are not included in the initial vocabulary, resulting in suboptimal exploration in chemical space.

\paragraph{Fragment remasking.}
Therefore, utilizing discrete diffusion of GenMol, we propose \emph{fragment remasking}, an effective strategy to explore the neighborhood of a given molecule in chemical space to find optimal molecules. Discrete diffusion allows GenMol to mask and re-predict some tokens of a given molecule, making neighborhood exploration simple and straightforward. However, although there exist some works that apply token-wise remasking to protein sequence optimization~\citep{hayes2024simulating,gruver2024protein}, it would be suboptimal to naively adopt the same strategy for small molecule optimization. This is because each token in a SAFE (or SMILES) sequence represents a single atom or bond, and masking them individually results in localized and ineffective exploration. The units that carry information about molecular properties are fragments or functional groups, not individual atoms or bonds~\citep{li2020application},
% Inspired by SAR, fragment remasking modifies molecules by randomly selecting a fragment in a given molecule and replacing it with a fragment mask chunk, from which discrete diffusion generates a new molecular fragment. We will examine the superiority of fragment remasking over token remasking in Section~\ref{sec:exp_ablation}.
and thus fragment remasking modifies a molecule by randomly re-predicting a fragment. 

Given $\pmb{x}_\text{init}$, $\pmb{x}_\text{mask}$ is constructed by randomly selecting one of the fragments of $\pmb{x}_\text{init}$ and replacing it with a mask chunk. Here, a finer decomposition rule than the one that constructed the vocabulary is used, allowing fragment remasking to operate at a fine-grained level.
GenMol then generates $\pmb{x}_\text{new}$ by iteratively unmasking $\pmb{x}_\text{mask}$. The proposed fragment remasking can also be viewed as a mutation operation in GA where the mutation is performed at the fragment-level rather than at the atom- or bond-level. $\pmb{x}_\text{new}$ is decomposed and scored by Eq.~(\ref{eq:score}), and the fragment vocabulary is dynamically updated by selecting the top-$V$ fragments, allowing exploration beyond the initial fragments. We summarize the goal-directed generation process of GenMol with fragment remasking in Algorithm~\ref{tab:algorithm}.

\begin{algorithm}[t!]
    \small
    \caption{\small Goal-directed Molecular Optimization of GenMol}
\begin{algorithmic}
    \STATE \textbf{Input:} A set of molecules $\mathcal{D}$, vocabulary size $V$, \\
    \quad\quad\quad decomposition rule for frag. vocabulary $R_\text{vocab}$, \\
    \quad\quad\quad decomposition rule for frag. remasking $R_\text{remask}$, \\
    \quad\quad\quad number of generations $G$ \\
    \STATE Set $\mathcal{F} \gets$ frags obtained by decomposing $\mathcal{D}$ with $R_\text{vocab}$ \\
    \STATE Set $\mathcal{V} \gets$ top-$V$ fragments of $\mathcal{F}$ (Eq.~\ref{eq:score}) \\
    \STATE Set $p_{\text{len}} \gets$ fragment length distribution of $\mathcal{D}$ \\
    \par\quad based on $R_\text{remask}$
    \STATE Set $\mathcal{M} \gets \emptyset$
    \WHILE{$\vert\mathcal{M}\vert < G$}
        \STATE Select and attach two fragments from $\mathcal{V}$ to get $\pmb{x}_\text{init}$
        \STATE Sample the fragment length $m \sim p_{\text{len}}$
        \STATE Select one of the fragments of $\pmb{x}_\text{init}$ based on $R_\text{remask}$
        \par\quad and replace it with $m$ mask tokens to get $\pmb{x}_\text{mask}$
        \STATE Generate $\pmb{x}_\text{new}$ by iteratively unmasking $\pmb{x}_\text{mask}$
        \STATE Update $\mathcal{M} \gets \mathcal{M} \cup \{\pmb{x}_\text{new}\}$
        \STATE Decompose $\pmb{x}_\text{new}$ into $\{\pmb{f}_1, \pmb{f}_2, \dots\}$ with $\mathcal{R}_\text{vocab}$
        \STATE Update $\mathcal{V} \gets$ top-$V$ fragments from $\mathcal{V} \cup \{\pmb{f}_1, \pmb{f}_2, \dots\}$
    \ENDWHILE
    \STATE \textbf{Output:} Generated molecules $\mathcal{M}$
\end{algorithmic}
\label{tab:algorithm}
\end{algorithm}

Instead of using a fixed length, the lengths of the fragment mask chunks are sampled from a predefined distribution, e.g., the distribution of fragment lengths in the training set. The length of the fragment depends on the decomposition rule used, and this strategy allows GenMol to automatically adjust the length based on the rule. This also ensures that GenMol generates fragments of varying lengths, offering users better controllability over molecule generation.

Fragment remasking can also be interpreted as Gibbs sampling~\citep{geman1984stochastic}. Assuming a SAFE molecular sequence $\pmb{x}$ is comprised of $F$ fragments, we can represent $\pmb{x}$ as a set of the fragments $\{\pmb{f}_k\}^F_{k=1}$, where $\pmb{f}_k$ denotes an attachment point-assigned SAFE fragment. To sample a molecule $\pmb{x}$ from $p(\pmb{x})=p(\pmb{f}_1, \dots, \pmb{f}_F)$, fragment remasking repeats the process of uniformly selecting the index $k$ and then sampling $\pmb{f}_k$ from $p(\pmb{f}_k|\pmb{f}_{\backslash k})$, where $\pmb{f}_{\backslash k}$ denotes $\pmb{f}_1, \dots, \pmb{f}_F$ but with $\pmb{f}_k$ omitted. This is equivalent to performing Gibbs sampling with the Markov kernel $p(\pmb{f}_k|\pmb{f}_{\backslash k})$, allowing GenMol to perform a random walk in the neighborhood of the given molecule $\pmb{x}$.
\looseness=-1

\begin{figure*}[t!]
    \begin{minipage}{0.71\linewidth}
        \centering
        \captionof{table}{\small \textbf{\emph{De novo} molecule generation results.} The results are the means and the standard deviations of 3 runs. $N$, $\tau$, and $r$ is the number of tokens to unmask at each time step, the softmax temperature, and the randomness, respectively. The best results are highlighted in bold.}
        \vspace{-0.1in}
        \centering
        \resizebox{\textwidth}{!}{
        \renewcommand{\arraystretch}{0.85}
        \renewcommand{\tabcolsep}{0.5mm}
        \begin{tabular}{lccccc}
        \Xhline{0.2ex}
            \rule{0pt}{10pt}Method & Validity (\%) & Uniqueness (\%) & \cellcolor{gray!25} Quality (\%) & Diversity & Sampling time (s) \\
        \hline
            \rule{0pt}{10pt}SAFE-GPT & \phantom{0}94.0~$\pm$~0.4 & \textbf{100.0}~$\pm$~0.0 & \cellcolor{gray!25} 54.7~$\pm$~0.3 & 0.879~$\pm$~0.001 & 27.7~$\pm$~0.1 \\
        \hline
            \rule{0pt}{10pt}GenMol w/o conf. sampling & \phantom{0}96.7~$\pm$~0.3 & \phantom{0}99.3~$\pm$~0.2 & \cellcolor{gray!25} 53.8~$\pm$~1.7 & 0.896~$\pm$~0.001 & 25.4~$\pm$~1.6 \\
        \hline
            \rule{0pt}{10pt}GenMol ($\tau=0.5,r=0.5$) & & & \cellcolor{gray!25} & & \\
            $\quad N=1$ & \textbf{100.0}~$\pm$~0.0 & \phantom{0}99.7~$\pm$~0.1 & \cellcolor{gray!25} \textbf{84.6}~$\pm$~0.8 & 0.818~$\pm$~0.001 & 21.1~$\pm$~0.4 \\
            $\quad N=2$ & \phantom{0}97.6~$\pm$~0.7 & \phantom{0}99.5~$\pm$~0.2 & \cellcolor{gray!25} 76.2~$\pm$~1.3 & 0.843~$\pm$~0.002 & 12.2~$\pm$~0.6 \\
            $\quad N=3$ & \phantom{0}95.6~$\pm$~0.5 & \phantom{0}99.0~$\pm$~0.1 & \cellcolor{gray!25} 67.1~$\pm$~0.7 & 0.861~$\pm$~0.001 & \textbf{10.1}~$\pm$~0.2 \\
        \hline
            \rule{0pt}{10pt}GenMol ($N=1$) & & & \cellcolor{gray!25} & & \\
            $\quad \tau=0.5,r=0.5$ & \textbf{100.0}~$\pm$~0.0 & \phantom{0}99.7~$\pm$~0.1 & \cellcolor{gray!25} \textbf{84.6}~$\pm$~0.8 & 0.818~$\pm$~0.001 & 21.1~$\pm$~0.4 \\
            $\quad \tau=0.5,r=1.0$ & \phantom{0}99.7~$\pm$~0.1 & \textbf{100.0}~$\pm$~0.0 & \cellcolor{gray!25} 83.8~$\pm$~0.5 & 0.832~$\pm$~0.001 & 20.5~$\pm$~0.6 \\
            % $\quad \tau=0.5,r=10.0$ & \phantom{0}99.6~$\pm$~0.1 & \phantom{0}99.9~$\pm$~0.0 & \cellcolor{gray!25} 75.0~$\pm$~0.6 & 0.858~$\pm$~0.002 & 21.6~$\pm$~0.4 \\
            $\quad \tau=1.0,r=1.0$ & \phantom{0}99.8~$\pm$~0.1 & \textbf{100.0}~$\pm$~0.1 & \cellcolor{gray!25} 79.1~$\pm$~0.9 & 0.845~$\pm$~0.002 & 21.9~$\pm$~0.6 \\
            $\quad \tau=1.0,r=10.0$ & \phantom{0}99.8~$\pm$~0.1 & \phantom{0}99.6~$\pm$~0.1 & \cellcolor{gray!25} 63.0~$\pm$~0.4 & 0.882~$\pm$~0.003 & 21.5~$\pm$~0.5 \\
            $\quad \tau=1.5,r=10.0$ & \phantom{0}95.6~$\pm$~0.3 & \phantom{0}98.3~$\pm$~0.2 & \cellcolor{gray!25} 39.7~$\pm$~0.5 & \textbf{0.911}~$\pm$~0.004 & 20.9~$\pm$~0.5 \\
        \Xhline{0.2ex}
        \end{tabular}}
        \label{tab:de_novo}
    \end{minipage}
    \hfill
    \begin{minipage}{0.28\linewidth}
        \centering
        \includegraphics[width=\textwidth]{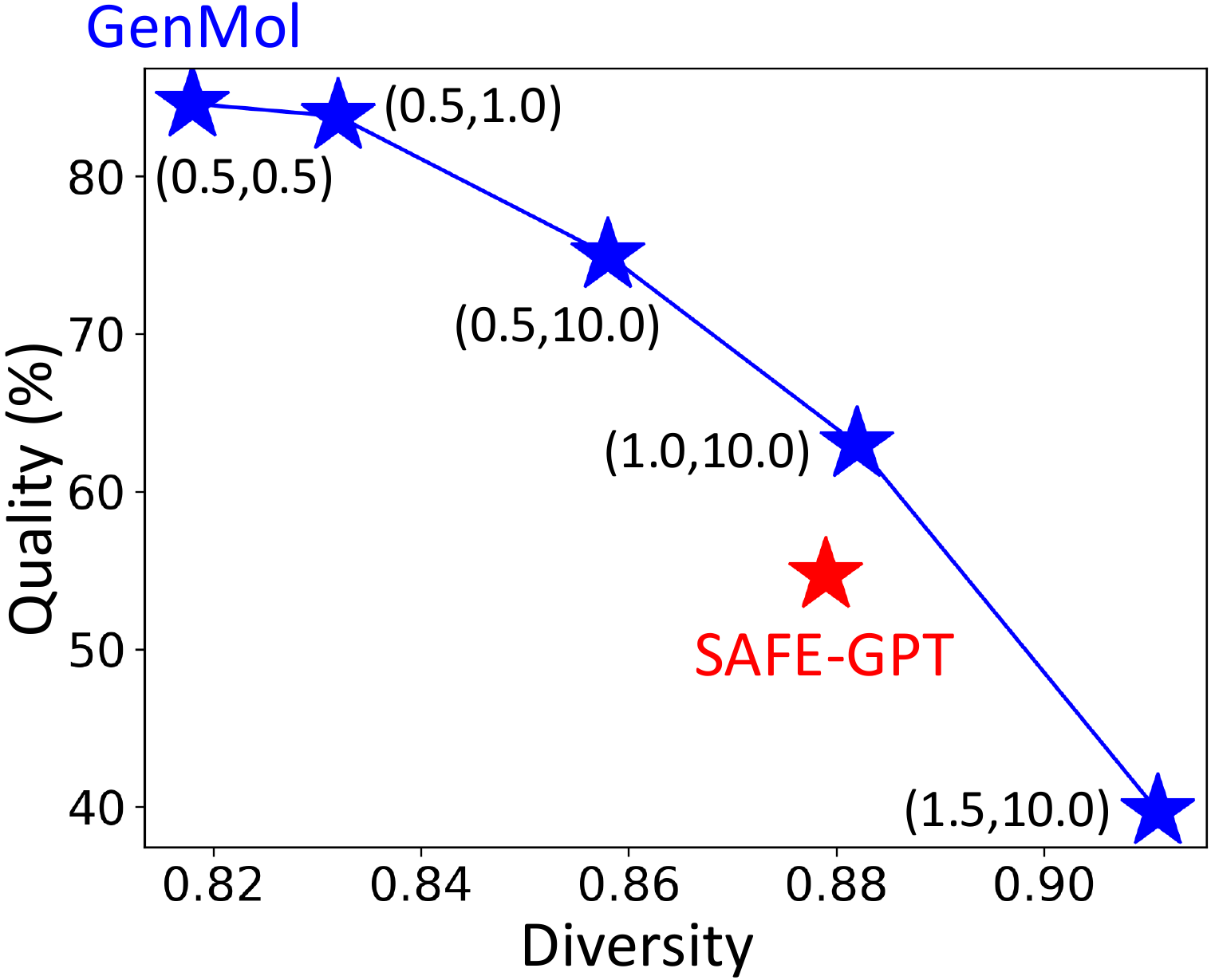}
        \vspace{-0.18in}
        \caption{\small \textbf{The quality-diversity trade-off in \emph{de novo} generation} with different values of $(\tau,r)$.}
        \label{fig:tradeoff}
    \end{minipage}
    % \vspace{-0.12in}
\end{figure*}

\subsection{Molecular Context Guidance~\label{sec:method_guidance}}

Inspired by autoguidance~\citep{karras2024guiding}, we propose \emph{molecular context guidance} (MCG), a guidance method tailored for masked discrete diffusion of GenMol. \citet{karras2024guiding} generalized classifier-free guidance (CFG)~\citep{ho2021classifier} and proposed to extrapolate between the predictions of two denoising networks:
\looseness=-1
\begin{equation}
    \fontsize{9pt}{9pt}
    D^{(w)}(\pmb{z}_s|\pmb{z}_t,\pmb{y}) = wD_1(\pmb{z}_s|\pmb{z}_t,\pmb{y})+(1-w)D_0(\pmb{z}_s|\pmb{z}_t,\pmb{y}),
    \label{eq:autoguidance}
\end{equation}
where $D_1$ and $D_0$ are a high-quality model and a poor model, respectively, $\pmb{y}$ is the condition to perform the guidance on, and $w>1$ is the guidance scale. Typically the same network $D_\theta$ is used for $D_1$ and $D_0$ with additional degradations applied to $D_0$, such as corrupted input, e.g., CFG sets $D_1=D_\theta(\pmb{z}_s|\pmb{z}_t,\pmb{y})$ and $D_0=D_\theta(\pmb{z}_s|\pmb{z}_t,\emptyset)$. The idea of autoguidance is that the weaker version of the same model amplifies the errors, and thus emphasizing the output of $D_1$ over $D_0$ by setting $w>1$ eliminates these errors.

\begin{table*}[t!]
    \caption{\small \textbf{Fragment-constrained molecule generation results.} The results are the means and the standard deviations of 3 runs. The best results are highlighted in bold.}
    \vspace{-0.1in}
    \centering
    \resizebox{0.9\textwidth}{!}{
    \renewcommand{\arraystretch}{0.88}
    \begin{tabular}{llccccc}
    \Xhline{0.2ex}
        \rule{0pt}{10pt}Method & Task & Validity (\%) & Uniqueness (\%) & \cellcolor{gray!25} Quality (\%) & Diversity & Distance \\
    \hline
        \rule{0pt}{10pt}SAFE-GPT & Linker design & \phantom{0}76.6~$\pm$~5.1 & 82.5~$\pm$~1.9 & \cellcolor{gray!25} 21.7~$\pm$~1.1 & 0.545~$\pm$~0.007 & 0.541~$\pm$~0.006 \\
        & Scaffold morphing & \phantom{0}58.9~$\pm$~6.8 & 70.4~$\pm$~5.7 & \cellcolor{gray!25} 16.7~$\pm$~2.3 & 0.514~$\pm$~0.011 & 0.528~$\pm$~0.009 \\
        & Motif extension & \phantom{0}\textbf{96.1}~$\pm$~1.9 & 66.8~$\pm$~1.2 & \cellcolor{gray!25} 18.6~$\pm$~2.1 & 0.562~$\pm$~0.003 & 0.665~$\pm$~0.006 \\
        & Scaffold decoration & \phantom{0}\textbf{97.7}~$\pm$~0.3 & 74.7~$\pm$~2.5 & \cellcolor{gray!25} 10.0~$\pm$~1.4 & 0.575~$\pm$~0.008 & 0.625~$\pm$~0.009 \\
        & Superstructure generation & \phantom{0}95.7~$\pm$~2.0 & 83.0~$\pm$~5.9 & \cellcolor{gray!25} 14.3~$\pm$~3.7 & 0.573~$\pm$~0.028 & \textbf{0.776}~$\pm$~0.036 \\
    \hline
        \rule{0pt}{10pt}GenMol & Linker design & \textbf{100.0}~$\pm$~0.0 & \textbf{83.7}~$\pm$~0.5 & \cellcolor{gray!25} \textbf{21.9}~$\pm$~0.4 & \textbf{0.547}~$\pm$~0.002 & \textbf{0.563}~$\pm$~0.003 \\
        & Scaffold morphing & \textbf{100.0}~$\pm$~0.0 & \textbf{83.7}~$\pm$~0.5 & \cellcolor{gray!25} \textbf{21.9}~$\pm$~0.4 & \textbf{0.547}~$\pm$~0.002 & \textbf{0.563}~$\pm$~0.003 \\
        & Motif extension & \phantom{0}82.9~$\pm$~0.1 & \textbf{77.5}~$\pm$~0.1 & \cellcolor{gray!25} \textbf{30.1}~$\pm$~0.4 & \textbf{0.617}~$\pm$~0.002 & \textbf{0.682}~$\pm$~0.001 \\
        & Scaffold decoration & \phantom{0}96.6~$\pm$~0.8 & \textbf{82.7}~$\pm$~1.8 & \cellcolor{gray!25} \textbf{31.8}~$\pm$~0.5 & \textbf{0.591}~$\pm$~0.001 & \textbf{0.651}~$\pm$~0.001 \\
        & Superstructure generation & \phantom{0}\textbf{97.5}~$\pm$~0.9 & \textbf{83.6}~$\pm$~1.0 & \cellcolor{gray!25} \textbf{34.8}~$\pm$~1.0 & \textbf{0.599}~$\pm$~0.009 & 0.762~$\pm$~0.007 \\
    \Xhline{0.2ex}
    \end{tabular}}
    \label{tab:frag}
    \vspace{-0.1in}
\end{table*}

However, the above guidance method has only been applied to continuous diffusion and its application to discrete diffusion has remained unexplored. On the other hand, \citet{nisonoff2024unlocking} proposed a CFG scheme for continuous-time Markov chains (CTMCs)~\citep{campbell2022continuous}, a subclass of discrete diffusion models based on a continuous-time formulation.
As GenMol is trained under the MDLM framework which can be interpreted as a CTMC~\citep{sahoo2024simple}, we introduce autoguidance for the MDLM formulation. Specifically, MCG sets $D_1$ as the original prediction of the denoiser $\pmb{x}_\theta$ and $D_0$ as the prediction with partially corrupted input, replacing the logits in Eq.~(\ref{eq:p_l_i}) by
\begin{align}
    \fontsize{8pt}{8pt}
    \hspace{-5mm}
    \log\pmb{x}^{(w),l}_{\theta,i}(\pmb{z}_t,\negthinspace t) \negthinspace:=\negthinspace w\log\pmb{x}^l_{\theta,i} (\pmb{z}_t,\negthinspace t) \negthinspace+\negthinspace (1 \negthinspace-\negthinspace w)\negthinspace \log\pmb{x}^l_{\theta,i}(\tilde{\pmb{z}}_t,\negthinspace t),
    \hspace{-3mm}
    \label{eq:genmol_guidance}
\end{align}
where $\tilde{\pmb{z}}_t$ is constructed by masking $\gamma\cdot100$\% of the tokens in $\pmb{z}_t$. We provide the derivation in Section~\ref{sec:derivation}. Intuitively, Eq.~(\ref{eq:genmol_guidance}) compares two outputs from a single GenMol model, with good and poor input, respectively. Specifically, the good input is a given partially masked sequence, which is further masked by $\gamma\cdot100$\% to yield poor input, and the two resulting logits are compared to calibrate GenMol's predictions. Using MCG, GenMol can fully utilize the given molecular context information in fragment-constrained and goal-directed generation.

\section{Experiments}

GenMol is trained on the SAFE dataset~\citep{noutahi2024gotta}, which combines molecules from ZINC~\citep{irwin2012zinc} and UniChem~\citep{chambers2013unichem}.
We emphasize that a single GenMol checkpoint is used to perform the following tasks without any additional finetuning specific to each task. We first conduct experiments on \emph{de novo} molecule generation in Section~\ref{sec:exp_de_novo}. Next, we conduct experiments on fragment-constrained molecule generation tasks in Section~\ref{sec:exp_frag}. We then examine GenMol's ability to perform goal-directed hit generation and goal-directed lead optimization in Section~\ref{sec:exp_pmo} and Section~\ref{sec:exp_lead}, respectively. We perform ablation studies in Section~\ref{sec:exp_ablation}.

\subsection{\emph{De Novo} Generation~\label{sec:exp_de_novo}}

\paragraph{Setup.}
In \emph{de novo} generation, the goal is to generate valid, unique, and diverse molecules.
We generate 1,000 molecules and evaluate them with the following metrics, following \citet{noutahi2024gotta}. \textbf{Validity} is the fraction of generated molecules that are chemically valid. \textbf{Uniqueness} is the fraction of valid molecules that are unique. \textbf{Diversity} is defined as the average pairwise Tanimoto distance between the Morgan fingerprints of the generated molecules. We further introduce \textbf{quality}, the fraction of valid, unique, drug-like, and synthesizable molecules, to provide a single metric that evaluates the ability to generate chemically reasonable and unique molecules. Here, \emph{drug-like} and \emph{synthesizable} molecules are defined as those satisfying quantitative estimate of drug-likeness (QED)~\citep{bickerton2012quantifying} $\ge 0.6$ and synthetic accessibility (SA)~\citep{ertl2009estimation} $\le 4$, respectively, following \citet{jin2020multi}. Further details are provided in Section~\ref{sec:detail_de_novo}.

\paragraph{Results.}
The results are shown in Table~\ref{tab:de_novo}. GenMol w/o conf. sampling is a GenMol that uses the standard diffusion sampling~\citep{austin2021structured,sahoo2024simple} instead of confidence sampling. SAFE-GPT~\citep{noutahi2024gotta}, GenMol w/o conf. sampling, and GenMol all show a near-perfect uniqueness, while GenMol significantly outperforms the other two in terms of validity, quality, and sampling time. Thanks to the non-autoregressive parallel decoding scheme, GenMol can predict multiple tokens simultaneously and shows much faster sampling as $N$, the number of tokens to unmask at each time step, increases. Notably, GenMol with $N=3$ shows higher quality than SAFE-GPT and GenMol w/o conf. sampling with 2.5x shorter sampling time and comparable diversity. Furthermore, GenMol can balance between quality and diversity by adjusting the values of the softmax temperature $\tau$ and the randomness $r$ of the confidence sampling. This balance is also shown in Figure~\ref{fig:tradeoff}, demonstrating that GenMol generates molecules along the Pareto frontier of the quality-diversity trade-off. Further analysis on quality and diversity are provided in Section~\ref{app:add_qed_sa}, Section~\ref{app:add_quality_diversity} and Section~\ref{app:add_quality}.
\looseness=-1

\begin{table*}[t!]
    \centering
    \caption{\small \textbf{Goal-directed hit generation results.} The results are the means of PMO AUC top-10 of 3 runs. The results for $f$-RAG~\citep{lee2024frag}, Genetic GFN~\citep{kim2024genetic} and Mol GA~\citep{tripp2023genetic} are taken from the respective papers and the results for other baselines are taken from \citet{gao2022sample}. The best results are highlighted in bold.}
    \vspace{-0.1in}
    \resizebox{0.9\textwidth}{!}{
    \renewcommand{\arraystretch}{0.88}
    \begin{tabular}{l|cccccccc}
    \Xhline{0.2ex}
        \rule{0pt}{10pt}Oracle & GenMol & $f$-RAG & Genetic GFN & Mol GA & REINVENT & Graph GA & \makecell{SELFIES-\\REINVENT} & GP BO \\
    \hline
        \rule{0pt}{10pt}albuterol\_similarity & 0.937 & \textbf{0.977} & 0.949 & 0.896 & 0.882 & 0.838 & 0.826 & 0.898 \\
        amlodipine\_mpo & \textbf{0.810} & 0.749 & 0.761 & 0.688 & 0.635 & 0.661 & 0.607 & 0.583 \\
        celecoxib\_rediscovery & \textbf{0.826} & 0.778 & 0.802 & 0.567 & 0.713 & 0.630 & 0.573 & 0.723 \\
        deco\_hop & \textbf{0.960} & 0.936 & 0.733 & 0.649 & 0.666 & 0.619 & 0.631 & 0.629 \\
        drd2 & \textbf{0.995} & 0.992 & 0.974 & 0.936 & 0.945 & 0.964 & 0.943 & 0.923 \\
        fexofenadine\_mpo & \textbf{0.894} & 0.856 & 0.856 & 0.825 & 0.784 & 0.760 & 0.741 & 0.722 \\
        gsk3b & \textbf{0.986} & 0.969 & 0.881 & 0.843 & 0.865 & 0.788 & 0.780 & 0.851 \\
        isomers\_c7h8n2o2 & 0.942 & 0.955 & \textbf{0.969} & 0.878 & 0.852 & 0.862 & 0.849 & 0.680 \\
        isomers\_c9h10n2o2pf2cl & 0.833 & 0.850 & \textbf{0.897} & 0.865 & 0.642 & 0.719 & 0.733 & 0.469 \\
        jnk3 & \textbf{0.906} & 0.904 & 0.764 & 0.702 & 0.783 & 0.553 & 0.631 & 0.564 \\
        median1 & \textbf{0.398} & 0.340 & 0.379 & 0.257 & 0.356 & 0.294 & 0.355 & 0.301 \\
        median2 & \textbf{0.359} & 0.323 & 0.294 & 0.301 & 0.276 & 0.273 & 0.255 & 0.297 \\
        mestranol\_similarity & \textbf{0.982} & 0.671 & 0.708 & 0.591 & 0.618 & 0.579 & 0.620 & 0.627 \\
        osimertinib\_mpo & \textbf{0.876} & 0.866 & 0.860 & 0.844 & 0.837 & 0.831 & 0.820 & 0.787 \\
        perindopril\_mpo & \textbf{0.718} & 0.681 & 0.595 & 0.547 & 0.537 & 0.538 & 0.517 & 0.493 \\
        qed & \textbf{0.942} & 0.939 & \textbf{0.942} & 0.941 & 0.941 & 0.940 & 0.940 & 0.937 \\
        ranolazine\_mpo & \textbf{0.821} & 0.820 & 0.819 & 0.804 & 0.760 & 0.728 & 0.748 & 0.735 \\
        scaffold\_hop & \textbf{0.628} & 0.576 & 0.615 & 0.527 & 0.560 & 0.517 & 0.525 & 0.548 \\
        sitagliptin\_mpo & 0.584 & 0.601 & \textbf{0.634} & 0.582 & 0.021 & 0.433 & 0.194 & 0.186 \\
        thiothixene\_rediscovery & \textbf{0.692} & 0.584 & 0.583 & 0.519 & 0.534 & 0.479 & 0.495 & 0.559 \\
        troglitazone\_rediscovery & \textbf{0.867} & 0.448 & 0.511 & 0.427 & 0.441 & 0.390 & 0.348 & 0.410 \\
        valsartan\_smarts & \textbf{0.822} & 0.627 & 0.135 & 0.000 & 0.178 & 0.000 & 0.000 & 0.000 \\
        zaleplon\_mpo & \textbf{0.584} & 0.486 & 0.552 & 0.519 & 0.358 & 0.346 & 0.333 & 0.221 \\
    \hline
        \rule{0pt}{10pt}Sum & \textbf{18.362} & 16.928 & 16.213 & 14.708 & 14.196 & 13.751 & 13.471 & 13.156 \\
    \Xhline{0.2ex}
    \end{tabular}}
    \label{tab:pmo}
    \vspace{-0.1in}
\end{table*}

\subsection{Fragment-constrained Generation~\label{sec:exp_frag}}

\paragraph{Setup.} In fragment-constrained generation, the goal is to complete a molecule given a set of fragments, a frequently encountered setting in real-world drug discovery. We use the benchmark of \citet{noutahi2024gotta}, which uses input fragments extracted from 10 known drugs to perform \textbf{linker design}, \textbf{scaffold morphing}, \textbf{motif extension}, \textbf{scaffold decoration}, and \textbf{superstructure generation}. In addition to \textbf{validity}, \textbf{uniqueness}, \textbf{diversity} and \textbf{quality} introduced in Section~\ref{sec:exp_de_novo}, \textbf{distance}, the average Tanimoto distance between the original and generated molecules, is also measured. 100 molecules are generated for each drug and averaged. Further details are provided in Section~\ref{sec:detail_frag}.
\looseness=-1

\paragraph{Results.}
The results of fragment-constrained generation are shown in Table~\ref{tab:frag}. GenMol significantly outperforms SAFE-GPT on most metrics across the tasks, demonstrating its general applicability to a variety of fragment constraint generation settings. Especially, GenMol generates high-quality molecules while preserving high diversity under the fragment constraints, again validating GenMol can strike an improved balance between quality and diversity.

\subsection{Goal-directed Hit Generation~\label{sec:exp_pmo}}

\vspace{-0.05in}
\paragraph{Setup.}
In goal-directed hit generation, the goal is to generate hits, i.e., molecules with optimized target chemical properties. Following \citet{lee2023drug} and \citet{lee2024frag}, we construct an initial fragment vocabulary by decomposing the molecules in the ZINC250k dataset~\citep{irwin2012zinc}. We adopt the practical molecular optimization (PMO) benchmark~\citep{gao2022sample} which contains 23 tasks. The maximum number of oracle calls is set to 10,000 and performance is measured using the area under the curve (AUC) of the average top-10 property scores versus oracle calls. As our baselines, we adopt the recent state-of-the-art methods, $f$-RAG~\citep{lee2024frag}, Genetic GFN~\citep{kim2024genetic}, and Mol GA~\citep{tripp2023genetic}. We also report the results of the top four methods in \citet{gao2022sample}. Note that since \citet{gao2022sample} reported the results of a total of 25 methods, comparing GenMol to the top methods is equivalent to comparing it to 25 methods. Further details are provided in Section~\ref{sec:detail_pmo}.

\vspace{-0.05in}
\paragraph{Results.}
The results are shown in Table~\ref{tab:pmo}. As shown in the table, GenMol significantly outperforms the previous methods in terms of the sum AUC top-10 value and exhibits the best performance in 19 out of 23 tasks by a large margin. These results verify that the proposed optimization strategy of GenMol with fragment remasking is effective in discovering optimized hits. The results of additional baselines are provided in Table~\ref{tab:app_pmo}, Table~\ref{tab:app_pmo_2}, and Table~\ref{tab:app_pmo_3}.

\subsection{Goal-directed Lead Optimization~\label{sec:exp_lead}}

\paragraph{Setup.}
Given an initial seed molecule, the goal in goal-directed lead optimization is to generate leads, i.e., molecules that exhibit improved target properties while maintaining the similarity with the given seed. Following \citet{wang2023retrieval}, the objective is to optimize the binding affinity to the target protein while satisfying the following constraints: $\text{QED}\ge0.6$, $\text{SA}\le4$, and $\text{sim}\ge\delta$ where $\delta\in\{0.4,0.6\}$ and sim is the Tanimoto similarity between the Morgan fingerprints of the generated molecules and the seed. Performance is evaluated by the docking score of the most optimized lead. Following \citet{lee2023exploring}, we adopt five target proteins, \textbf{parp1}, \textbf{fa7}, \textbf{5ht1b}, \textbf{braf}, and \textbf{jak2}. For each target, three molecules from its known active ligands are selected and each is given as a seed molecule, yielding a total of 30 tasks. An initial fragment vocabulary is constructed by decomposing the seed molecule. If the generated molecule is lead, its fragments are added to the vocabulary. Following \citet{wang2023retrieval}, we adopt Graph GA~\citep{jensen2019graph} and RetMol~\citep{wang2023retrieval} as our baselines. Further details are provided in Section~\ref{sec:detail_lead}.

\vspace{-0.05in}
\paragraph{Results.}
The results of goal-directed lead optimization are shown in Table~\ref{tab:lead}. As shown in the table, GenMol outperforms the baselines in most tasks. Note that baselines often fail, i.e., they cannot generate molecules with a higher binding affinity than the seed molecule while satisfying the constraints, especially under the harsher ($\delta=0.6$) similarity constraint. In contrast, GenMol is able to successfully optimize seed molecules in 26 out of 30 tasks, validating its effectiveness in exploring chemical space to optimize given molecules and discover promising lead molecules.

\begin{table*}[t!]
    \centering
    \begin{minipage}{0.58\linewidth}
        \caption{\small \textbf{Lead optimization results (kcal/mol).} The results are the mean docking scores of the most optimized leads of 3 runs. Lower is better.}
        \vspace{-0.1in}
        \centering
        \resizebox{\textwidth}{!}{
        \renewcommand{\arraystretch}{0.8}
        \renewcommand{\tabcolsep}{1.5mm}
        \begin{tabular}{cc|ccc|ccc}
        \Xhline{0.2ex}
            \rule{0pt}{10pt}\multirow{2}{*}{\makecell{Target\\protein}} & \multirow{2}{*}{\makecell{Seed\\score}} & \multicolumn{3}{c|}{$\delta=0.4$} & \multicolumn{3}{c}{$\delta=0.6$} \\
            \Xcline{3-8}{0.1ex} & & \rule{0pt}{10pt}GenMol & RetMol & Graph GA & GenMol & RetMol & Graph GA \\
        \hline
            \rule{0pt}{10pt} & -7.3 & \textbf{-10.6} & \phantom{0}-9.0 & \phantom{0}-8.3 & \textbf{-10.4} & - & \phantom{0}-8.6 \\
            parp1 & -7.8 & \textbf{-11.0} & -10.7 & \phantom{0}-8.9 & \phantom{0}\textbf{-9.7} & - & \phantom{0}-8.1 \\
            & -8.2 & \textbf{-11.3} & -10.9 & - & \phantom{0}\textbf{-9.2} & - & - \\
        \hline
            \rule{0pt}{10pt} & -6.4 & \phantom{0}\textbf{-8.4} & \phantom{0}-8.0 & \phantom{0}-7.8 & \phantom{0}-7.3 & \phantom{0}\textbf{-7.6} & \phantom{0}\textbf{-7.6} \\
            fa7 & -6.7 & \phantom{0}\textbf{-8.4} & - & \phantom{0}-8.2 & \phantom{0}\textbf{-7.6} & - & \phantom{0}\textbf{-7.6} \\
            & -8.5 & - & - & - & - & - & - \\
        \hline
            \rule{0pt}{10pt} & -4.5 & \textbf{-12.9} & -12.1 & -11.7 & \textbf{-12.1} & - & -11.3 \\
            5ht1b & -7.6 & \textbf{-12.3} & \phantom{0}-9.0 & -12.1 & \textbf{-12.0} & -10.0 & \textbf{-12.0} \\
            & -9.8 & \textbf{-11.6} & - & - & \textbf{-10.5} & - & - \\
        \hline
            \rule{0pt}{10pt} & -9.3 & \textbf{-10.8} & - & \phantom{0}-9.8 & - & - & - \\
            braf & -9.4 & -10.8 & \textbf{-11.6} & - & \phantom{0}\textbf{-9.7} & - & - \\
            & -9.8 & -10.6 & - & \textbf{-11.6} & \textbf{-10.5} & - & -10.4 \\
        \hline
            \rule{0pt}{10pt} & -7.7 & \textbf{-10.2} & \phantom{0}-8.2 & \phantom{0}-8.7 & \phantom{0}\textbf{-9.3} & - & \phantom{0}-8.1 \\
            jak2 & -8.0 & \textbf{-10.0} & \phantom{0}-9.0 & \phantom{0}-9.2 & \phantom{0}\textbf{-9.4} & - & \phantom{0}-9.2 \\
            & -8.6 & \phantom{0}\textbf{-9.8} & - & - & - & - & - \\
        \Xhline{0.2ex}
        \end{tabular}}
        \label{tab:lead}
    \end{minipage}
    \quad
    \begin{minipage}{0.32\linewidth}
        \caption{\small \textbf{Ablation study} on the goal-directed hit generation task. The results are the mean sums of PMO AUC top-10 of 3 runs. The best results are highlighted in bold. The full results are shown in Table~\ref{tab:app_pmo_ablation}.}
        \vspace{0.1in}
        \resizebox{\textwidth}{!}{
        \renewcommand{\arraystretch}{0.8}
        \renewcommand{\tabcolsep}{1mm}
        \begin{tabular}{l|c}
        \Xhline{0.2ex}
            \rule{0pt}{10pt}Method & Sum \\
        \hline
            \rule{0pt}{10pt}Attaching (A) & 17.641 \\
            \\
            \rule{0pt}{10pt}A + Token remasking & 18.091 \\
            & \textcolor{violet}{\scriptsize{(A+0.450)}} \\
            \rule{0pt}{10pt}A + GPT remasking & 18.074 \\
            & \textcolor{violet}{\scriptsize{(A+0.433)}} \\
            \rule{0pt}{10pt}A + Fragment remasking (F) & 18.208 \\
            & \textcolor{violet}{\scriptsize{(A+0.567)}} \\
            \rule{0pt}{10pt}A + F + MCG & \textbf{18.362} \\
            & \textcolor{violet}{\scriptsize{(A+F+0.154)}} \\
        \Xhline{0.2ex}
        \end{tabular}}
        \label{tab:pmo_ablation}
    \end{minipage}
    \vspace{-0.1in}
\end{table*}

\vspace{-0.05in}
\subsection{Ablation Study~\label{sec:exp_ablation}}

\vspace{-0.05in}
\paragraph{Fragment remasking.}
To examine the effect of the proposed fragment remasking with masked discrete diffusion, we conduct ablation studies with alternative remasking strategies in Table~\ref{tab:pmo_ablation}. \textbf{Attaching (A)} is a baseline that attaches two fragments from the vocabulary without further modifications. On top of it, \textbf{A + Token remasking} randomly re-predicts individual tokens instead of a fragment chunk with discrete diffusion and \textbf{A + GPT remasking} re-predicts a randomly chosen fragment chunk with SAFE-GPT instead of diffusion. \textbf{A + Fragment remasking (F)} re-predicts a fragment mask chunk with discrete diffusion. First, A + Token remasking, A + GPT remasking and A + Fragment remasking all outperform A, highlighting the importance of exploration through remasking. A + Fragment remasking outperforms A + Token remasking, proving that using fragments as the exploration unit is aligned with chemical intuition and effective in chemical exploration. A + Fragment remasking is also superior to A + GPT remasking, proving the effectiveness of the masked discrete diffusion that does not rely on specific ordering of tokens with bidirectional attention.
We also conduct the ablation studies on lead optimization in Table~\ref{tab:app_lead_ablation}. Although the naive baseline A outperforms other previous baselines in hit generation, it cannot generate new fragments outside of the vocabulary and therefore fails frequently in lead optimization, and applying fragment remasking on top of it largely improves lead optimization performance.

\vspace{-0.05in}
\paragraph{Molecular context guidance.}
To verify the effect of MCG, we present the results of GenMol with (\textbf{A + F + MCG}) and without (\textbf{A + F}; i.e., $\gamma=0$) MCG in Table~\ref{tab:pmo_ablation}. A + F + MCG shows its superiority over A + F, demonstrating that calibrating GenMol's predictions with molecular context information with MCG improves GenMol's performance. The full results are shown in Table~\ref{tab:app_pmo_ablation}, where A + F + MCG achieves the best performance in 19 out of 23 tasks. The same trend is also observed in Table~\ref{tab:app_frag}, where GenMol w/ MCG outperforms GenMol w/o MCG across various tasks on fragment-constrained generation.
\looseness=-1

\section{Conclusion}

We proposed GenMol, a molecule generation framework designed to deal with various drug discovery scenarios effectively and efficiently by integrating discrete diffusion with SAFE. Especially, fragment remasking allows GenMol to effectively explore chemical space and MCG further improves GenMol's performance. The experimental results showed that GenMol can achieve state-of-the-art results in a wide range of drug discovery tasks, demonstrating its potential as a unified and versatile tool for drug discovery.

\section*{Impact Statement}
In our paper, we showed that GenMol is capable of addressing a broad spectrum of drug discovery challenges, providing a unified and versatile solution for molecular design. However, as effective as GenMol is in drug discovery tasks, it has the potential to generate harmful drugs if used maliciously. To prevent this, GenMol could be equipped with features that incorporate target properties that take toxicity into account, exclude toxic fragments from the fragment vocabulary, or filter the proposed drug candidates by predicting the toxicity.

\clearpage
{
    \small
    \bibliography{references}

\begin{thebibliography}{57}
\providecommand{\natexlab}[1]{#1}
\providecommand{\url}[1]{\texttt{#1}}
\expandafter\ifx\csname urlstyle\endcsname\relax
  \providecommand{\doi}[1]{doi: #1}\else
  \providecommand{\doi}{doi: \begingroup \urlstyle{rm}\Url}\fi

\bibitem[Austin et~al.(2021)Austin, Johnson, Ho, Tarlow, and Van Den~Berg]{austin2021structured}
Austin, J., Johnson, D.~D., Ho, J., Tarlow, D., and Van Den~Berg, R.
\newblock Structured denoising diffusion models in discrete state-spaces.
\newblock \emph{Advances in Neural Information Processing Systems}, 34:\penalty0 17981--17993, 2021.

\bibitem[Bickerton et~al.(2012)Bickerton, Paolini, Besnard, Muresan, and Hopkins]{bickerton2012quantifying}
Bickerton, G.~R., Paolini, G.~V., Besnard, J., Muresan, S., and Hopkins, A.~L.
\newblock Quantifying the chemical beauty of drugs.
\newblock \emph{Nature chemistry}, 4\penalty0 (2):\penalty0 90--98, 2012.

\bibitem[Brown et~al.(2019)Brown, Fiscato, Segler, and Vaucher]{brown2019guacamol}
Brown, N., Fiscato, M., Segler, M.~H., and Vaucher, A.~C.
\newblock Guacamol: benchmarking models for de novo molecular design.
\newblock \emph{Journal of chemical information and modeling}, 59\penalty0 (3):\penalty0 1096--1108, 2019.

\bibitem[Campbell et~al.(2022)Campbell, Benton, De~Bortoli, Rainforth, Deligiannidis, and Doucet]{campbell2022continuous}
Campbell, A., Benton, J., De~Bortoli, V., Rainforth, T., Deligiannidis, G., and Doucet, A.
\newblock A continuous time framework for discrete denoising models.
\newblock \emph{Advances in Neural Information Processing Systems}, 35:\penalty0 28266--28279, 2022.

\bibitem[Chambers et~al.(2013)Chambers, Davies, Gaulton, Hersey, Velankar, Petryszak, Hastings, Bellis, McGlinchey, and Overington]{chambers2013unichem}
Chambers, J., Davies, M., Gaulton, A., Hersey, A., Velankar, S., Petryszak, R., Hastings, J., Bellis, L., McGlinchey, S., and Overington, J.~P.
\newblock Unichem: a unified chemical structure cross-referencing and identifier tracking system.
\newblock \emph{Journal of cheminformatics}, 5\penalty0 (1):\penalty0 3, 2013.

\bibitem[Chang et~al.(2022)Chang, Zhang, Jiang, Liu, and Freeman]{chang2022maskgit}
Chang, H., Zhang, H., Jiang, L., Liu, C., and Freeman, W.~T.
\newblock Maskgit: Masked generative image transformer.
\newblock In \emph{Proceedings of the IEEE/CVF Conference on Computer Vision and Pattern Recognition}, pp.\  11315--11325, 2022.

\bibitem[Degen et~al.(2008)Degen, Wegscheid-Gerlach, Zaliani, and Rarey]{brics}
Degen, J., Wegscheid-Gerlach, C., Zaliani, A., and Rarey, M.
\newblock On the art of compiling and using 'drug-like' chemical fragment spaces.
\newblock \emph{ChemMedChem: Chemistry Enabling Drug Discovery}, 3\penalty0 (10):\penalty0 1503--1507, 2008.

\bibitem[Devlin et~al.(2019)Devlin, Chang, Lee, and Toutanova]{devlin2018bert}
Devlin, J., Chang, M.-W., Lee, K., and Toutanova, K.
\newblock Bert: Pre-training of deep bidirectional transformers for language understanding.
\newblock In \emph{Proceedings of naacL-HLT}, 2019.

\bibitem[Ertl \& Schuffenhauer(2009)Ertl and Schuffenhauer]{ertl2009estimation}
Ertl, P. and Schuffenhauer, A.
\newblock Estimation of synthetic accessibility score of drug-like molecules based on molecular complexity and fragment contributions.
\newblock \emph{Journal of cheminformatics}, 1:\penalty0 1--11, 2009.

\bibitem[Gao et~al.(2022)Gao, Fu, Sun, and Coley]{gao2022sample}
Gao, W., Fu, T., Sun, J., and Coley, C.
\newblock Sample efficiency matters: a benchmark for practical molecular optimization.
\newblock \emph{Advances in Neural Information Processing Systems}, 35:\penalty0 21342--21357, 2022.

\bibitem[Geman \& Geman(1984)Geman and Geman]{geman1984stochastic}
Geman, S. and Geman, D.
\newblock Stochastic relaxation, gibbs distributions, and the bayesian restoration of images.
\newblock \emph{IEEE Transactions on pattern analysis and machine intelligence}, pp.\  721--741, 1984.

\bibitem[Geng et~al.(2023)Geng, Xie, Xia, Wu, Qin, Wang, Zhang, Wu, and Liu]{geng2023novo}
Geng, Z., Xie, S., Xia, Y., Wu, L., Qin, T., Wang, J., Zhang, Y., Wu, F., and Liu, T.-Y.
\newblock De novo molecular generation via connection-aware motif mining.
\newblock In \emph{International Conference on Learning Representations}, 2023.

\bibitem[Gruver et~al.(2024)Gruver, Stanton, Frey, Rudner, Hotzel, Lafrance-Vanasse, Rajpal, Cho, and Wilson]{gruver2024protein}
Gruver, N., Stanton, S., Frey, N., Rudner, T.~G., Hotzel, I., Lafrance-Vanasse, J., Rajpal, A., Cho, K., and Wilson, A.~G.
\newblock Protein design with guided discrete diffusion.
\newblock \emph{Advances in neural information processing systems}, 36, 2024.

\bibitem[Guo et~al.(2023)Guo, Knuth, Margreitter, Janet, Papadopoulos, Engkvist, and Patronov]{guo2023link}
Guo, J., Knuth, F., Margreitter, C., Janet, J.~P., Papadopoulos, K., Engkvist, O., and Patronov, A.
\newblock Link-invent: generative linker design with reinforcement learning.
\newblock \emph{Digital Discovery}, 2\penalty0 (2):\penalty0 392--408, 2023.

\bibitem[Hayes et~al.(2024)Hayes, Rao, Akin, Sofroniew, Oktay, Lin, Verkuil, Tran, Deaton, Wiggert, et~al.]{hayes2024simulating}
Hayes, T., Rao, R., Akin, H., Sofroniew, N.~J., Oktay, D., Lin, Z., Verkuil, R., Tran, V.~Q., Deaton, J., Wiggert, M., et~al.
\newblock Simulating 500 million years of evolution with a language model.
\newblock \emph{bioRxiv}, pp.\  2024--07, 2024.

\bibitem[He et~al.(2023)He, Sun, Tang, Wang, Huang, and Qiu]{he2022diffusionbert}
He, Z., Sun, T., Tang, Q., Wang, K., Huang, X., and Qiu, X.
\newblock Diffusionbert: Improving generative masked language models with diffusion models.
\newblock In \emph{The 61st Annual Meeting Of The Association For Computational Linguistics}, 2023.

\bibitem[Ho \& Salimans(2021)Ho and Salimans]{ho2021classifier}
Ho, J. and Salimans, T.
\newblock Classifier-free diffusion guidance.
\newblock In \emph{NeurIPS 2021 Workshop on Deep Generative Models and Downstream Applications}, 2021.

\bibitem[Hoogeboom et~al.(2021)Hoogeboom, Nielsen, Jaini, Forr{\'e}, and Welling]{hoogeboom2021argmax}
Hoogeboom, E., Nielsen, D., Jaini, P., Forr{\'e}, P., and Welling, M.
\newblock Argmax flows and multinomial diffusion: Learning categorical distributions.
\newblock \emph{Advances in Neural Information Processing Systems}, 34:\penalty0 12454--12465, 2021.

\bibitem[Hua et~al.(2024)Hua, Luan, Xu, Ying, Fu, Ermon, and Precup]{hua2024mudiff}
Hua, C., Luan, S., Xu, M., Ying, Z., Fu, J., Ermon, S., and Precup, D.
\newblock Mudiff: Unified diffusion for complete molecule generation.
\newblock In \emph{Learning on Graphs Conference}, pp.\  33--1. PMLR, 2024.

\bibitem[Huang et~al.(2021)Huang, Fu, Gao, Zhao, Roohani, Leskovec, Coley, Xiao, Sun, and Zitnik]{huang2021therapeutics}
Huang, K., Fu, T., Gao, W., Zhao, Y., Roohani, Y.~H., Leskovec, J., Coley, C.~W., Xiao, C., Sun, J., and Zitnik, M.
\newblock Therapeutics data commons: Machine learning datasets and tasks for drug discovery and development.
\newblock In \emph{NeurIPS Track Datasets and Benchmarks}, 2021.

\bibitem[Hughes et~al.(2011)Hughes, Rees, Kalindjian, and Philpott]{hughes2011principles}
Hughes, J.~P., Rees, S., Kalindjian, S.~B., and Philpott, K.~L.
\newblock Principles of early drug discovery.
\newblock \emph{British journal of pharmacology}, 162\penalty0 (6):\penalty0 1239--1249, 2011.

\bibitem[Irwin et~al.(2012)Irwin, Sterling, Mysinger, Bolstad, and Coleman]{irwin2012zinc}
Irwin, J.~J., Sterling, T., Mysinger, M.~M., Bolstad, E.~S., and Coleman, R.~G.
\newblock Zinc: a free tool to discover chemistry for biology.
\newblock \emph{Journal of chemical information and modeling}, 52\penalty0 (7):\penalty0 1757--1768, 2012.

\bibitem[Jensen(2019)]{jensen2019graph}
Jensen, J.~H.
\newblock A graph-based genetic algorithm and generative model/monte carlo tree search for the exploration of chemical space.
\newblock \emph{Chemical science}, 10\penalty0 (12):\penalty0 3567--3572, 2019.

\bibitem[Jin et~al.(2018)Jin, Barzilay, and Jaakkola]{jin2018junction}
Jin, W., Barzilay, R., and Jaakkola, T.
\newblock Junction tree variational autoencoder for molecular graph generation.
\newblock In \emph{International Conference on Machine Learning}, pp.\  2323--2332. PMLR, 2018.

\bibitem[Jin et~al.(2020)Jin, Barzilay, and Jaakkola]{jin2020multi}
Jin, W., Barzilay, R., and Jaakkola, T.
\newblock Multi-objective molecule generation using interpretable substructures.
\newblock In \emph{International conference on machine learning}, pp.\  4849--4859. PMLR, 2020.

\bibitem[Karras et~al.(2024)Karras, Aittala, Kynk{\"a}{\"a}nniemi, Lehtinen, Aila, and Laine]{karras2024guiding}
Karras, T., Aittala, M., Kynk{\"a}{\"a}nniemi, T., Lehtinen, J., Aila, T., and Laine, S.
\newblock Guiding a diffusion model with a bad version of itself.
\newblock \emph{Advances in Neural Information Processing Systems}, 2024.

\bibitem[Kim et~al.(2024)Kim, Kim, Choi, and Park]{kim2024genetic}
Kim, H., Kim, M., Choi, S., and Park, J.
\newblock Genetic-guided gflownets: Advancing in practical molecular optimization benchmark.
\newblock \emph{Advances in Neural Information Processing Systems}, 2024.

\bibitem[Kong et~al.(2022)Kong, Huang, Tan, and Liu]{kong2022molecule}
Kong, X., Huang, W., Tan, Z., and Liu, Y.
\newblock Molecule generation by principal subgraph mining and assembling.
\newblock \emph{Advances in Neural Information Processing Systems}, 35:\penalty0 2550--2563, 2022.

\bibitem[Landrum et~al.(2016)]{landrum2016rdkit}
Landrum, G. et~al.
\newblock {RDKit}: Open-source cheminformatics software, 2016.
\newblock \emph{URL http://www. rdkit. org/, https://github. com/rdkit/rdkit}, 2016.

\bibitem[Lee et~al.(2023)Lee, Jo, and Hwang]{lee2023exploring}
Lee, S., Jo, J., and Hwang, S.~J.
\newblock Exploring chemical space with score-based out-of-distribution generation.
\newblock In \emph{International Conference on Machine Learning}, pp.\  18872--18892. PMLR, 2023.

\bibitem[Lee et~al.(2024{\natexlab{a}})Lee, Kreis, Veccham, Liu, Reidenbach, Paliwal, Vahdat, and Nie]{lee2024frag}
Lee, S., Kreis, K., Veccham, S.~P., Liu, M., Reidenbach, D., Paliwal, S., Vahdat, A., and Nie, W.
\newblock Molecule generation with fragment retrieval augmentation.
\newblock \emph{Advances in Neural Information Processing Systems}, 2024{\natexlab{a}}.

\bibitem[Lee et~al.(2024{\natexlab{b}})Lee, Lee, Kawaguchi, and Hwang]{lee2023drug}
Lee, S., Lee, S., Kawaguchi, K., and Hwang, S.~J.
\newblock Drug discovery with dynamic goal-aware fragments.
\newblock \emph{International Conference on Machine Learning}, 2024{\natexlab{b}}.

\bibitem[Li(2020)]{li2020application}
Li, Q.
\newblock Application of fragment-based drug discovery to versatile targets.
\newblock \emph{Frontiers in molecular biosciences}, 7:\penalty0 180, 2020.

\bibitem[Lin et~al.(2024)Lin, Huang, Zhang, Ma, Liu, Li, Wu, Ji, Hou, and Li]{lin2024diffbp}
Lin, H., Huang, Y., Zhang, O., Ma, S., Liu, M., Li, X., Wu, L., Ji, S., Hou, T., and Li, S.~Z.
\newblock Diffbp: Generative diffusion of 3d molecules for target protein binding.
\newblock \emph{Chemical Science}, 2024.

\bibitem[Loshchilov \& Hutter(2019)Loshchilov and Hutter]{loshchilov2017decoupled}
Loshchilov, I. and Hutter, F.
\newblock Decoupled weight decay regularization.
\newblock \emph{International Conference on Learning Representations}, 2019.

\bibitem[Lou et~al.(2024)Lou, Meng, and Ermon]{lou2023discrete}
Lou, A., Meng, C., and Ermon, S.
\newblock Discrete diffusion language modeling by estimating the ratios of the data distribution.
\newblock \emph{International Conference on Machine Learning}, 2024.

\bibitem[Maziarz et~al.(2021)Maziarz, Jackson-Flux, Cameron, Sirockin, Schneider, Stiefl, Segler, and Brockschmidt]{maziarz2021learning}
Maziarz, K., Jackson-Flux, H.~R., Cameron, P., Sirockin, F., Schneider, N., Stiefl, N., Segler, M., and Brockschmidt, M.
\newblock Learning to extend molecular scaffolds with structural motifs.
\newblock In \emph{International Conference on Learning Representations}, 2021.

\bibitem[Murray \& Rees(2009)Murray and Rees]{murray2009rise}
Murray, C.~W. and Rees, D.~C.
\newblock The rise of fragment-based drug discovery.
\newblock \emph{Nature chemistry}, 1\penalty0 (3):\penalty0 187--192, 2009.

\bibitem[Mysinger et~al.(2012)Mysinger, Carchia, Irwin, and Shoichet]{mysinger2012directory}
Mysinger, M.~M., Carchia, M., Irwin, J.~J., and Shoichet, B.~K.
\newblock Directory of useful decoys, enhanced (dud-e): better ligands and decoys for better benchmarking.
\newblock \emph{Journal of medicinal chemistry}, 55\penalty0 (14):\penalty0 6582--6594, 2012.

\bibitem[Nisonoff et~al.(2024)Nisonoff, Xiong, Allenspach, and Listgarten]{nisonoff2024unlocking}
Nisonoff, H., Xiong, J., Allenspach, S., and Listgarten, J.
\newblock Unlocking guidance for discrete state-space diffusion and flow models.
\newblock \emph{arXiv preprint arXiv:2406.01572}, 2024.

\bibitem[Noutahi et~al.(2024)Noutahi, Gabellini, Craig, Lim, and Tossou]{noutahi2024gotta}
Noutahi, E., Gabellini, C., Craig, M., Lim, J.~S., and Tossou, P.
\newblock Gotta be safe: a new framework for molecular design.
\newblock \emph{Digital Discovery}, 3\penalty0 (4):\penalty0 796--804, 2024.

\bibitem[Olivecrona et~al.(2017)Olivecrona, Blaschke, Engkvist, and Chen]{olivecrona2017molecular}
Olivecrona, M., Blaschke, T., Engkvist, O., and Chen, H.
\newblock Molecular de-novo design through deep reinforcement learning.
\newblock \emph{Journal of cheminformatics}, 9\penalty0 (1):\penalty0 1--14, 2017.

\bibitem[Powers et~al.(2023)Powers, Yu, Suriana, Koodli, Lu, Paggi, and Dror]{powers2023geometric}
Powers, A.~S., Yu, H.~H., Suriana, P., Koodli, R.~V., Lu, T., Paggi, J.~M., and Dror, R.~O.
\newblock Geometric deep learning for structure-based ligand design.
\newblock \emph{ACS Central Science}, 9\penalty0 (12):\penalty0 2257--2267, 2023.

\bibitem[Sahoo et~al.(2024)Sahoo, Arriola, Schiff, Gokaslan, Marroquin, Chiu, Rush, and Kuleshov]{sahoo2024simple}
Sahoo, S.~S., Arriola, M., Schiff, Y., Gokaslan, A., Marroquin, E., Chiu, J.~T., Rush, A., and Kuleshov, V.
\newblock Simple and effective masked diffusion language models.
\newblock \emph{Advances in Neural Information Processing Systems}, 2024.

\bibitem[Shi et~al.(2024)Shi, Han, Wang, Doucet, and Titsias]{shi2024simplified}
Shi, J., Han, K., Wang, Z., Doucet, A., and Titsias, M.~K.
\newblock Simplified and generalized masked diffusion for discrete data.
\newblock \emph{Advances in neural information processing systems}, 2024.

\bibitem[Sohl-Dickstein et~al.(2015)Sohl-Dickstein, Weiss, Maheswaranathan, and Ganguli]{sohl2015deep}
Sohl-Dickstein, J., Weiss, E., Maheswaranathan, N., and Ganguli, S.
\newblock Deep unsupervised learning using nonequilibrium thermodynamics.
\newblock In \emph{International conference on machine learning}, pp.\  2256--2265. PMLR, 2015.

\bibitem[Tripp \& Hern{\'a}ndez-Lobato(2023)Tripp and Hern{\'a}ndez-Lobato]{tripp2023genetic}
Tripp, A. and Hern{\'a}ndez-Lobato, J.~M.
\newblock Genetic algorithms are strong baselines for molecule generation.
\newblock \emph{arXiv preprint arXiv:2310.09267}, 2023.

\bibitem[Vignac et~al.(2023)Vignac, Krawczuk, Siraudin, Wang, Cevher, and Frossard]{vignac2022digress}
Vignac, C., Krawczuk, I., Siraudin, A., Wang, B., Cevher, V., and Frossard, P.
\newblock Digress: Discrete denoising diffusion for graph generation.
\newblock In \emph{Proceedings of the 11th International Conference on Learning Representations}, 2023.

\bibitem[Wang et~al.(2023)Wang, Nie, Qiao, Xiao, Baraniuk, and Anandkumar]{wang2023retrieval}
Wang, Z., Nie, W., Qiao, Z., Xiao, C., Baraniuk, R., and Anandkumar, A.
\newblock Retrieval-based controllable molecule generation.
\newblock In \emph{International Conference on Learning Representations}, 2023.

\bibitem[Weininger(1988)]{weininger1988smiles}
Weininger, D.
\newblock Smiles, a chemical language and information system. 1. introduction to methodology and encoding rules.
\newblock \emph{Journal of chemical information and computer sciences}, 28\penalty0 (1):\penalty0 31--36, 1988.

\bibitem[Wolf et~al.(2019)Wolf, Debut, Sanh, Chaumond, Delangue, Moi, Cistac, Rault, Louf, Funtowicz, et~al.]{wolf2019huggingface}
Wolf, T., Debut, L., Sanh, V., Chaumond, J., Delangue, C., Moi, A., Cistac, P., Rault, T., Louf, R., Funtowicz, M., et~al.
\newblock Huggingface’s transformers: State-of-the-art natural language processing. arxiv.
\newblock \emph{arXiv preprint arXiv:1910.03771}, 2019.

\bibitem[Xie et~al.(2020)Xie, Shi, Zhou, Yang, Zhang, Yu, and Li]{xie2020mars}
Xie, Y., Shi, C., Zhou, H., Yang, Y., Zhang, W., Yu, Y., and Li, L.
\newblock Mars: Markov molecular sampling for multi-objective drug discovery.
\newblock In \emph{International Conference on Learning Representations}, 2020.

\bibitem[Yang et~al.(2021)Yang, Hwang, Lee, Ryu, and Hwang]{yang2021hit}
Yang, S., Hwang, D., Lee, S., Ryu, S., and Hwang, S.~J.
\newblock Hit and lead discovery with explorative rl and fragment-based molecule generation.
\newblock \emph{Advances in Neural Information Processing Systems}, 34:\penalty0 7924--7936, 2021.

\bibitem[Yang et~al.(2020)Yang, Zheng, Su, Zhao, Xu, and Chen]{yang2020syntalinker}
Yang, Y., Zheng, S., Su, S., Zhao, C., Xu, J., and Chen, H.
\newblock Syntalinker: automatic fragment linking with deep conditional transformer neural networks.
\newblock \emph{Chemical science}, 11\penalty0 (31):\penalty0 8312--8322, 2020.

\bibitem[Zdrazil et~al.(2024)Zdrazil, Felix, Hunter, Manners, Blackshaw, Corbett, de~Veij, Ioannidis, Lopez, Mosquera, et~al.]{zdrazil2024chembl}
Zdrazil, B., Felix, E., Hunter, F., Manners, E.~J., Blackshaw, J., Corbett, S., de~Veij, M., Ioannidis, H., Lopez, D.~M., Mosquera, J.~F., et~al.
\newblock The chembl database in 2023: a drug discovery platform spanning multiple bioactivity data types and time periods.
\newblock \emph{Nucleic acids research}, 52\penalty0 (D1):\penalty0 D1180--D1192, 2024.

\bibitem[Zhang et~al.(2023)Zhang, Wang, Smith, Eaton, Rees, and Gu]{zhang2023diffmol}
Zhang, W., Wang, X., Smith, J., Eaton, J., Rees, B., and Gu, Q.
\newblock Diffmol: 3d structured molecule generation with discrete denoising diffusion probabilistic models.
\newblock In \emph{ICML 2023 Workshop on Structured Probabilistic Inference $\{$$\backslash$\&$\}$ Generative Modeling}, 2023.

\bibitem[Zheng et~al.(2024)Zheng, Yuan, Yu, and Kong]{zheng2023reparameterized}
Zheng, L., Yuan, J., Yu, L., and Kong, L.
\newblock A reparameterized discrete diffusion model for text generation.
\newblock \emph{Conference on Language Modeling}, 2024.

\end{thebibliography}
    \bibliographystyle{icml2025}
}
\appendix
\onecolumn

\begin{figure*}[t!]
    \centering
    \includegraphics[width=0.8\linewidth]{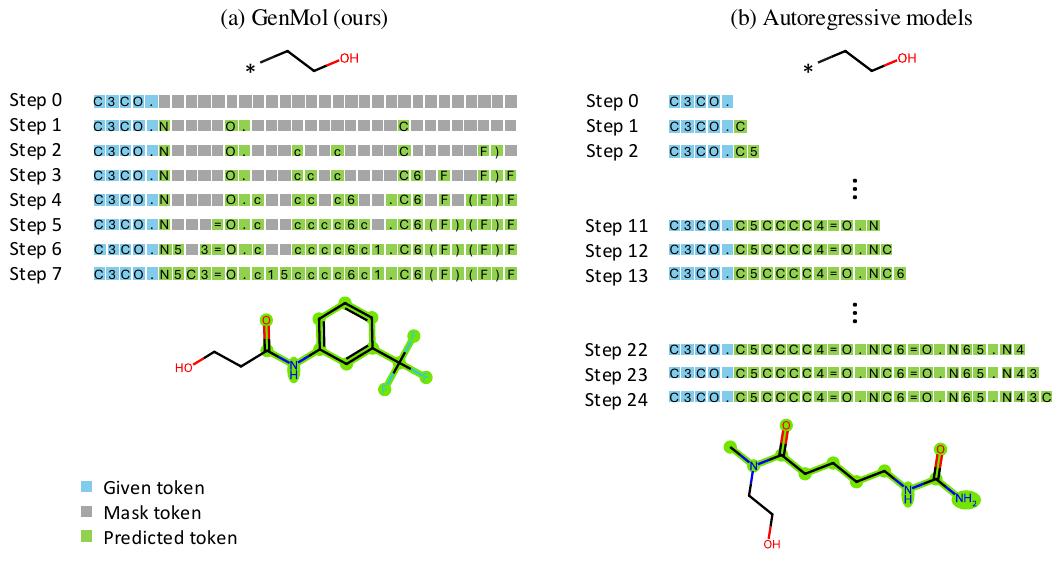}
    \caption{\small \textbf{Illustration of the fragment-constrained motif extension process of (a) GenMol and (b) autoregressive models.} GenMol starts by sampling the length of the sequence and then filling the sequence with mask tokens corresponding to the sampled length. GenMol employs parallel decoding where all tokens are decoded simultaneously under discrete diffusion, and confirms only the most confident predictions. The decoding proceeds progressively until all mask tokens are predicted. In contrast, autoregressive models like SAFE-GPT~\citep{noutahi2024gotta} need to predict one token per step, requiring many more decoding steps.}
    \label{fig:autoregressive}
\end{figure*}

\section{Limitations}
The proposed GenMol is designed with the goal of being a versatile tool for various drug discovery scenarios without any task-specific finetuning. In addition to its versatility, GenMol shows improved molecular generation in terms of both effectiveness and efficiency thanks to its masked discrete diffusion with non-autoregressive bidirectional parallel decoding. However, while the parallel decoding scheme reduces sampling time, unmasking more than one token at each time step (i.e., setting $N>1$) results in degraded generation quality. Overcoming this trade-off between generation quality and sampling efficiency is left as a future work.

\section{Confidence Sampling of GenMol~\label{sec:confidence_sampling}}

Following \citet{chang2022maskgit}, GenMol adopts confidence sampling that decides which tokens to unmask at each sampling step based on the confidence scores of the sampled tokens. After sampling $\pmb{z}^l_s$ according to Eq.~(\ref{eq:p_l_i}), let $p^l_{\theta,i^*}$ denote the corresponding prediction score, where $i^*$ is the index of the sampled category. With introduction of a Gumbel noise decreasing over the sampling process, the confidence score $c^l_t$ of the $l$-th token at time step $t$ is defined as follows:
\begin{align}
    c^l_t := \log p^l_{\theta,i^*} + r \cdot t \cdot \epsilon, \quad \epsilon \sim \text{Gumbel}(0,1),
    \label{eq:confidence}
\end{align}
where $r$ is the randomness.

Based on their confidence scores $c^l_t$ (Eq.~(\ref{eq:confidence})), GenMol unmask the top-$N$ currently masked tokens. In other words, this is equivalent to predicting all tokens simultaneously, but only confirming the most confident predictions. The other tokens are masked again and predicted in the next step. With confidence sampling, GenMol can exploit the dependencies between tokens in a given molecular sequence for better sampling quality, rather than randomly and independently selecting tokens to unmask, as in the standard diffusion sampling~\citep{austin2021structured,sahoo2024simple}.

\clearpage
\section{Derivation of MCG (Eq.~(\ref{eq:genmol_guidance}))~\label{sec:derivation}}

\citet{nisonoff2024unlocking} proposed a CFG scheme for CTMC-based discrete diffusion models~\citep{campbell2022continuous} as follows:
\begin{equation}
    R^{(w)}_t(\pmb{z}_t,\pmb{z}_s|\pmb{y})=R_t(\pmb{z}_t,\pmb{z}_s|\pmb{y})^wR_t(\pmb{z}_t,\pmb{z}_s)^{1-w},
    \label{eq:ctmc_guidance}
\end{equation}
where $R(\pmb{z},\pmb{z}')$ is a rate matrix that describes the transition probability for $\pmb{z}\rightarrow\pmb{z}'$.

Applying the same generalization of CFG proposed by \citet{karras2024guiding} (Eq.~(\ref{eq:autoguidance})) to Eq.~(\ref{eq:ctmc_guidance}), we have the following guided rate matrix:
\begin{equation}
    R^{(w)}_t(\pmb{z}_t,\pmb{z}_s|\pmb{y})=R_{1,t}(\pmb{z}_t,\pmb{z}_s|\pmb{y})^wR_{0,t}(\pmb{z}_t,\pmb{z}_s|\pmb{y})^{1-w},
    \label{eq:ctmc_autoguidance}
\end{equation}
where $R_{1,t}$ and $R_{0,t}$ are the rate matrices describing a high-quality model and a poor model, respectively.

On the other hand, MDLM can be interpreted as a CTMC with the following reverse rate matrix~\citep{sahoo2024simple}:
\begin{align}
    R_t(\pmb{z}_t,\pmb{z}_s)=
    \begin{cases}
    -\frac{\alpha'_t}{1-\alpha_t}\langle\pmb{x}_\theta(\pmb{z}_t,t),\pmb{z}_s\rangle &\pmb{z}_s\neq \textbf{m}, \pmb{z}_t=\textbf{m} \\
    \frac{\alpha'_t}{1-\alpha_t} &\pmb{z}_s=\textbf{m}, \pmb{z}_t=\textbf{m} \\
    0 &\text{otherwise}.
    \end{cases}
    \label{eq:mdlm_ctmc}
\end{align}

If we set $R_{1,t}$ and $R_{0,t}$ as the rate matrices resulted by the original prediction of the denoiser $\pmb{x}_\theta(\pmb{z}_t,t)$ and the prediction with partially corrupted input $\pmb{x}_\theta(\tilde{\pmb{z}}_t,t)$, respectively, substituting Eq.~(\ref{eq:mdlm_ctmc}) for $R_{1,t}$ and $R_{0,t}$ in Eq.~(\ref{eq:ctmc_autoguidance}) yields:
\begin{align}
\begin{split}
    R^{(w)}_t(\pmb{z}_t,\pmb{z}_s) &=
    \begin{cases}
    -\frac{\alpha'_t}{1-\alpha_t}\langle\pmb{x}_\theta(\pmb{z}_t,t),\pmb{z}_s\rangle^w \langle\pmb{x}_\theta(\tilde{\pmb{z}}_t,t),\pmb{z}_s\rangle^{1-w} &\pmb{z}_s\neq \textbf{m}, \pmb{z}_t=\textbf{m} \\
    \frac{\alpha'_t}{1-\alpha_t} &\pmb{z}_s=\textbf{m}, \pmb{z}_t=\textbf{m} \\
    0 &\text{otherwise}
    \end{cases} \\
    &= \begin{cases}
    -\frac{\alpha'_t}{1-\alpha_t}\langle\pmb{x}_\theta(\pmb{z}_t,t)^w\odot\pmb{x}_\theta(\tilde{\pmb{z}}_t,t)^{1-w},\pmb{z}_s\rangle \quad\quad\quad &\pmb{z}_s\neq \textbf{m}, \pmb{z}_t=\textbf{m} \\
    \frac{\alpha'_t}{1-\alpha_t} &\pmb{z}_s=\textbf{m}, \pmb{z}_t=\textbf{m} \\
    0 &\text{otherwise},
    \end{cases}
    \label{eq:genmol_ctmc}
\end{split}
\end{align}
where $\odot$ denotes the Hadamard product. Here, we utilized the facts that $\pmb{z}_s$ is an one-hot vector and $\pmb{z}_t=\textbf{m}\Rightarrow \tilde{\pmb{z}}_t=\textbf{m}$ as $\tilde{\pmb{z}}_t$ is the corrupted (i.e., more masked) $\pmb{z}_t$.

Therefore, using the guided rate matrix $R^{(w)}_t$ in Eq.~(\ref{eq:genmol_ctmc}) is equivalent to using the following guided prediction $\pmb{x}^{(w)}_\theta$:
\begin{align}
\begin{split}
    \pmb{x}^{(w)}_\theta(\pmb{z}_t,t) &= \pmb{x}_\theta(\pmb{z}_t,t)^w \odot \pmb{x}_\theta(\tilde{\pmb{z}}_t,t)^{1-w} \\
    \Leftrightarrow \,\, \log\pmb{x}^{(w)}_\theta(\pmb{z}_t,t) &= w\log\pmb{x}_\theta(\pmb{z}_t,t) + (1-w)\log\pmb{x}_\theta(\tilde{\pmb{z}}_t,t).
\end{split}
\end{align}

\clearpage
\section{Experimental Details}

\subsection{Computing Resources}
GenMol was trained using 8 NVIDIA A100 GPUs. The training took approximately 5 hours. All the molecular generation experiments were conducted using a single NVIDIA A100 GPU and 32 CPU cores.

\subsection{Training GenMol}

\begin{table*}[t!]
    \caption{\small \textbf{Statistics of the SAFE dataset.}}
    \centering
    \resizebox{0.6\textwidth}{!}{
    \renewcommand{\arraystretch}{0.95}
    \renewcommand{\tabcolsep}{2.5mm}
    \begin{tabular}{lccc}
    \Xhline{0.2ex}
        \rule{0pt}{10pt} & Train & Test & Validation \\
    \hline
        \rule{0pt}{10pt}Number of examples & 945,455,307 & 118,890,444 & 118,451,032 \\
    \Xhline{0.2ex}
    \end{tabular}}
    \label{tab:data}
\end{table*}

In this section, we describe the details for training GenMol. 
We used the BERT~\citep{devlin2018bert} architecture of the HuggingFace Transformers library~\citep{wolf2019huggingface} with the default configuration, except that we set max\_position\_embeddings to 256. We used the SAFE dataset and SAFE tokenizer~\citep{noutahi2024gotta} that has a vocabulary size of $K=1880$. The statistics of the dataset is provided in Table~\ref{tab:data}. We set the batch size to 2048, the learning rate to $3e-4$, and the number of training steps to 50k. We used the log-linear noise schedule of \citet{sahoo2024simple} and the AdamW optimizer~\citep{loshchilov2017decoupled} with $\beta_1=0.9$ and $\beta_2=0.999$.

\subsection{\emph{De Novo} Generation~\label{sec:detail_de_novo}}

In this section, we describe the details for conducting experiments in Section~\ref{sec:exp_de_novo}. We used the RDKit library~\citep{landrum2016rdkit} to obtain Morgan fingerprints and the Therapeutics Data Commons (TDC) library~\citep{huang2021therapeutics} to calculate diversity, QED, and SA. The lengths of the mask chunks were sampled from the ZINC250k distribution.

\subsection{Fragment-constrained Generation~\label{sec:detail_frag}}

In Section~\ref{sec:exp_frag}, we used the benchmark proposed by \citet{noutahi2024gotta}. The benchmark contains extracted fragments from 10 known drugs: Cyclothiazide, Maribavir, Spirapril, Baricitinib, Eliglustat, Erlotinib, Futibatinib, Lesinurad, Liothyronine, and Lovastatin. Specifically, from each drug, side chains, a starting motif, the main scaffold with attachment points, and a core substructure are extracted, and then serve as input for linker design \& scaffold morphing, motif extension, scaffold decoration, and superstructure generation, respectively. \textbf{Linker design} and \textbf{scaffold morphing} are tasks where the goal is to generate a linker fragment that connects given two side chains. In GenMol, linker design and scaffold morphing correspond to the same task. \textbf{Motif extension} and \textbf{scaffold decoration} are tasks where the goal is to generate a side fragment to complete a new molecule when a motif or scaffold and attachment points are given. \textbf{Superstructure generation} is a task where the goal is to generate a molecule when a substructure constraint is given. Following \citet{noutahi2024gotta}, we first generate random attachment points on the substructure to create new scaffolds and conduct the scaffold decoration task.

We used $N=1$. We performed the grid search with the search space $\tau \in \{0.5, 0.8, 1, 1.2, 1.5\}$ and $r \in \{1, 1.2, 2, 3\}$, and set $r$ to 3 for linker design and scaffold morphing, 1.2 for motif extension, and 2 for scaffold decoration and superstructure generation. We set $\tau$ to 1.2 for all the tasks. The lengths of the mask chunks were sampled from the ZINC250k distribution. For MCG, we set $w=2$ and performed a search with the search space $\gamma \in \{0, 0.1, 0.2, 0.3, 0.4, 0.5\}$. The values of $\gamma$ are provided in Table~\ref{tab:app_c_frag}.

\subsection{Goal-directed Hit Generation~\label{sec:detail_pmo}}

In this section, we describe the details for conducting experiments in Section~\ref{sec:exp_pmo}. To construct an initial fragment vocabulary, we adopted a simple decomposition rule $R_\text{vocab}$ that randomly cut one of the non-ring single bonds of a given molecule three times and apply it to the ZINC250k dataset. With this decomposition rule, we can ensure that all fragments have one attachment point and are of appropriate size. For the finer decomposition rule $R_\text{remask}$ that determines which fragments fragment remasking will operate on, we used the rule that cut all of the non-ring single bonds in a given molecule. We set the size of the fragment vocabulary to $V=100$. We applied the warmup scheme that let GenMol generate molecules by concatenating two randomly chosen fragments without fragment remasking for the first 1,000 generations. We used $N=1$, $\tau=1.2$, and $r=2$. For MCG, we set $w=2$ and performed a search with the search space $\gamma \in \{0, 0.1, 0.2, 0.3, 0.4, 0.5\}$. The values of $\gamma$ are shown in Table~\ref{tab:app_c_hit}.

\begin{table*}[t!]
    \begin{minipage}{0.28\linewidth}
        \caption{\small \textbf{$\gamma$ in fragment-constrained generation.}}
        \centering
        \resizebox{\textwidth}{!}{
        \begin{tabular}{lc}
        \Xhline{0.2ex}
            \rule{0pt}{10pt}Task & $\gamma$ \\
        \hline
            \rule{0pt}{10pt}Linker design & 0.0 \\
            Scaffold morphing & 0.0 \\
            Motif extension & 0.3 \\
            Scaffold decoration & 0.3 \\
            Superstructure generation & 0.4 \\
        \Xhline{0.2ex}
        \end{tabular}}
        \label{tab:app_c_frag}
    \end{minipage}
    \hfill
    \begin{minipage}{0.28\linewidth}
        \caption{\small \textbf{$\gamma$ in hit generation.}}
        \centering
        \resizebox{\textwidth}{!}{
        \renewcommand{\arraystretch}{0.95}
        \begin{tabular}{lc}
        \Xhline{0.2ex}
            \rule{0pt}{10pt}Oracle & $\gamma$ \\
        \hline
            \rule{0pt}{10pt}albuterol\_similarity & 0.2 \\
            amlodipine\_mpo & 0.3 \\
            celecoxib\_rediscovery & 0.0 \\
            deco\_hop & 0.2 \\
            drd2 & 0.0 \\
            fexofenadine\_mpo & 0.0 \\
            gsk3b & 0.0 \\
            isomers\_c7h8n2o2 & 0.5 \\
            isomers\_c9h10n2o2pf2cl & 0.0 \\
            jnk3 & 0.5 \\
            median1 & 0.2 \\
            median2 & 0.2 \\
            mestranol\_similarity & 0.0 \\
            osimertinib\_mpo & 0.0 \\
            perindopril\_mpo & 0.4 \\
            qed & 0.0 \\
            ranolazine\_mpo & 0.0 \\
            scaffold\_hop & 0.0 \\
            sitagliptin\_mpo & 0.2 \\
            thiothixene\_rediscovery & 0.3 \\
            troglitazone\_rediscovery & 0.0 \\
            valsartan\_smarts & 0.4 \\
            zaleplon\_mpo & 0.4 \\
        \Xhline{0.2ex}
        \end{tabular}}
        \label{tab:app_c_hit}
    \end{minipage}
    \hfill
    \begin{minipage}{0.33\linewidth}
        \caption{\small \textbf{$\gamma$ in lead optimization.}}
        \centering
        \resizebox{\textwidth}{!}{
        \renewcommand{\tabcolsep}{3mm}
        \begin{tabular}{cc|c}
        \Xhline{0.2ex}
            \rule{0pt}{10pt}Target protein & Seed score & $\gamma$ \\
        \hline
            \rule{0pt}{10pt} & -7.3 & 0.2 \\
            parp1 & -7.8 & 0.2 \\
            & -8.2 & 0.2 \\
        \hline
            \rule{0pt}{10pt} & -6.4 & 0.3 \\
            fa7 & -6.7 & 0.4 \\
            & -8.5 & 0.0 \\
        \hline
            \rule{0pt}{10pt} & -4.5 & 0.3 \\
            5ht1b & -7.6 & 0.0 \\
            & -9.8 & 0.4 \\
        \hline
            \rule{0pt}{10pt} & -9.3 & 0.2 \\
            braf & -9.4 & 0.1 \\
            & -9.8 & 0.5 \\
        \hline
            \rule{0pt}{10pt} & -7.7 & 0.5 \\
            jak2 & -8.0 & 0.0 \\
            & -8.6 & 0.1 \\
        \Xhline{0.2ex}
        \end{tabular}}
        \label{tab:app_c_lead}
    \end{minipage}
    \vspace{-0.1in}
\end{table*}

\subsection{Goal-directed Lead Optimization~\label{sec:detail_lead}}

In this section, we describe the details for conducting experiments in Section~\ref{sec:exp_lead}. For each target, three molecules were randomly selected from known active compounds from DUD-E~\citep{mysinger2012directory} (parp1, fa7, braf and jak2) or ChEMBL~\citep{zdrazil2024chembl} (5ht1b) as seed molecules. The same decomposition rules $R_\text{vocab}$ and $R_\text{remask}$ explained in Section~\ref{sec:detail_pmo} were used and the fragment vocabulary size was set to $V=\infty$. Following the setting of \citet{wang2023retrieval}, for each target protein and each seed molecule, we run 10 optimization iterations with 100 generation per iteration. We used $N=1$, $\tau=1.2$, and $r=2$. For MCG, we set $w=2$ and performed a search with the search space $\gamma \in \{0, 0.1, 0.2, 0.3, 0.4, 0.5\}$. The values of $\gamma$ are shown in Table~\ref{tab:app_c_lead}.

\section{Additional Experimental Results}

\subsection{Additional Baseline in \emph{De Novo} Generation~\label{app:add_de_novo}}
We provide comparison of GenMol with another widely used baseline trained on ZINC250k~\citep{irwin2012zinc}, JT-VAE~\citep{jin2018junction}, and a graph discrete diffusion model trained on GuacaMol~\citep{brown2019guacamol}, DiGress~\citep{vignac2022digress}, in Table~\ref{tab:app_de_novo_add}. GenMol significantly outperforms JT-VAE and DiGress in terms of quality and sampling time. We used the fast version of the JT-VAE code\footnote{\url{https://github.com/Bibyutatsu/FastJTNNpy3}} and the official code of DiGress\footnote{\url{https://github.com/cvignac/DiGress}}.

\subsection{QED and SA Distributions in \emph{De Novo} Generation~\label{app:add_qed_sa}}
\begin{figure}[t!]
    \centering
    \includegraphics[width=0.9\linewidth]{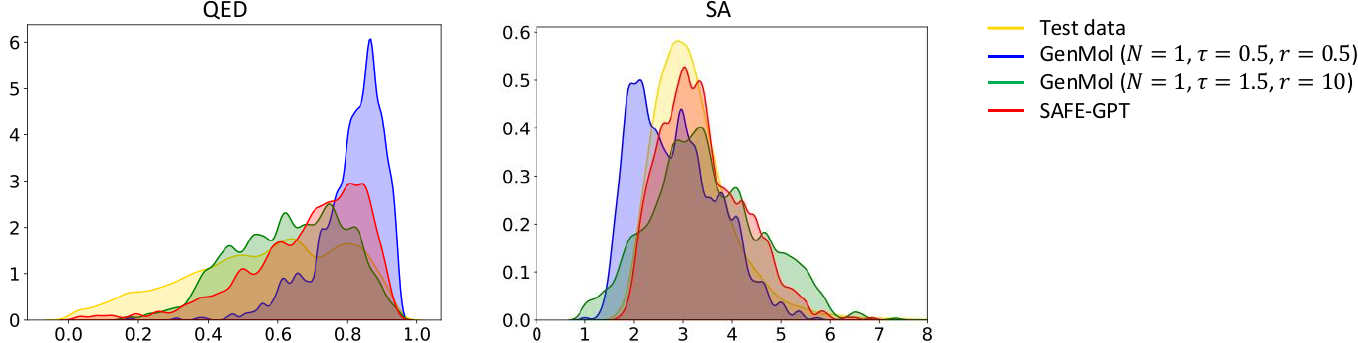}
    \caption{\small \textbf{QED and SA distributions of molecules in \emph{de novo} generation.}}
    \label{fig:qed_sa}
\end{figure}

We provide the QED and SA distributions of molecules generated by GenMol and SAFE-GPT, respectively, in Figure~\ref{fig:qed_sa}. The distributions of 100k molecules randomly sampled from the test set are also shown in the figure. As shown in the figure, GenMol is able to generate molecules of higher QED (more drug-like) and lower SA (more synthesizable) values than SAFE-GPT, resulting in high quality in Table~\ref{tab:de_novo}. Furthermore, GenMol can freely control these distributions by adjusting the values of the softmax temperature $\tau$ and the randomness $r$.

\subsection{Analysis on Quality and Diversity in \emph{De Novo} Generation~\label{app:add_quality_diversity}}
The quality and diversity values of 100k molecules randomly sampled from the test set are 38.2\% and 0.897, respectively. The quality and diversity values of molecules generated by GenMol ($N=1$, $\tau=0.5$, $r=0.5$) are 84.6\% and 0.818, respectively, as shown in Table~\ref{tab:de_novo}. We can think of this as sacrificing diversity and selecting a specific mode of high quality. GenMol can control this mode-selecting behavior by adjusting $\tau$ and $r$ (e.g., the quality and diversity values of GenMol ($N=1$, $\tau=1.5$, $r=10$) shown in Table~\ref{tab:de_novo} are 39.7\% and 0.911, respectively).

\subsection{Analysis on QED and SA Thresholds in Quality Metric~\label{app:add_quality}}
In Section~\ref{sec:exp_de_novo} and Section~\ref{sec:exp_frag}, we introduced quality to provide a metric that summarizes the model's ability to generate chemically plausible molecules. We followed \citet{jin2020multi} and set the QED and SA thresholds to 0.6 and 4, respectively. To validate the robustness of the quality metric to threshold selection, we additionally provide quality results on the \emph{de novo} generation task with different thresholds in Table~\ref{tab:app_de_novo}. $\text{QED}\ge0.5,\text{SA}\le5$ corresponds to the softer condition, while $\text{QED}\ge0.7,\text{SA}\le3$ is the harsher condition. As shown in the table, different QED and SA thresholds yield a consistent quality trend, with GenMol outperforming SAFE-GPT in all three settings.

\begin{table*}[t!]
    \caption{\small \textbf{\emph{De novo} molecule generation results.} The results are the means and the standard deviations of 3 runs. $N$, $\tau$, and $r$ is the number of tokens to unmask at each time step, the softmax temperature, and the randomness, respectively. The best results are highlighted in bold.}
    \vspace{-0.1in}
    \centering
    \resizebox{0.85\textwidth}{!}{
    \begin{tabular}{lccccc}
    \Xhline{0.2ex}
        \rule{0pt}{10pt}Method & Validity (\%) & Uniqueness (\%) & \cellcolor{gray!25} Quality (\%) & Diversity & Sampling time (s) \\
    \hline
        \rule{0pt}{10pt}JT-VAE & \textbf{100.0}~$\pm$~0.0 & \phantom{0}65.9~$\pm$~1.3 & \cellcolor{gray!25} 45.2~$\pm$~1.4 & 0.855~$\pm$~0.001 & \phantom{00}96.5~$\pm$~2.1 \\
        DiGress & \phantom{0}89.6~$\pm$~0.8 & \textbf{100.0}~$\pm$~0.0 & \cellcolor{gray!25} 36.8~$\pm$~1.0 & \textbf{0.885}~$\pm$~0.002 & 1241.9~$\pm$~9.2 \\
    \hline
        \rule{0pt}{10pt}GenMol ($N=1$) & & & \cellcolor{gray!25} & & \\
        $\quad \tau=0.5,r=0.5$ & \textbf{100.0}~$\pm$~0.0 & \phantom{0}99.7~$\pm$~0.1 & \cellcolor{gray!25} \textbf{84.6}~$\pm$~0.8 & 0.818~$\pm$~0.001 & \phantom{00}\textbf{21.1}~$\pm$~0.4 \\
        $\quad \tau=1.0,r=10.0$ & \phantom{0}99.8~$\pm$~0.1 & \phantom{0}99.6~$\pm$~0.1 & \cellcolor{gray!25} 63.0~$\pm$~0.4 & 0.882~$\pm$~0.003 & \phantom{00}21.5~$\pm$~0.5 \\
    \Xhline{0.2ex}
    \end{tabular}}
    \label{tab:app_de_novo_add}
\end{table*}

\begin{table*}[t!]
    \caption{\small \textbf{Quality (\%) results in \emph{de novo} molecule generation} with various QED and SA thresholds. The results are the means and the standard deviations of 3 runs. The best results are highlighted in bold.}
    \centering
    \resizebox{0.85\textwidth}{!}{
    \renewcommand{\arraystretch}{0.95}
    \renewcommand{\tabcolsep}{2.5mm}
    \begin{tabular}{lccc}
    \Xhline{0.2ex}
        \rule{0pt}{10pt}\multirow{2.2}{*}{Method} & \multicolumn{3}{c}{Quality} \\
        \Xcline{2-4}{0.1ex} & \rule{0pt}{10pt}$\text{QED}\ge0.6,\text{SA}\le4$ & $\text{QED}\ge0.5,\text{SA}\le5$ & $\text{QED}\ge0.7,\text{SA}\le3$ \\
    \hline
        \rule{0pt}{10pt}SAFE-GPT & 54.7~$\pm$~0.3 & 78.3~$\pm$~1.7 & 18.5~$\pm$~0.1 \\
        \rule{0pt}{10pt}GenMol ($N=1,\tau=0.5,r=0.5$) & \textbf{84.6}~$\pm$~0.8 & \textbf{97.1}~$\pm$~0.1 & \textbf{50.6}~$\pm$~1.0 \\
    \Xhline{0.2ex}
    \end{tabular}}
    \label{tab:app_de_novo}
\end{table*}

\subsection{Additional Baseline in Fragment-constrained Generation}
We compare the results of GenMol and DiGress in the fragment-constrained generation tasks in Table~\ref{tab:app_frag}. As shown in the table, GenMol significantly outperforms DiGress in validity and quality.

\subsection{Ablation Study on MCG in Fragment-constrained Generation~\label{app:add_frag}}
We compare the results of GenMol and GenMol without MCG in the fragment-constrained generation tasks in Table~\ref{tab:app_frag}. As shown in the table, using MCG improves the performance of GenMol across various tasks and metrics, verifying the effectiveness of the proposed MCG scheme.

\begin{table*}[t!]
    \caption{\small \textbf{Fragment-constrained molecule generation results.} The results are the means and the standard deviations of 3 runs. The \textcolor{violet}{purple parentheses} indicate the performance gain by MCG.}
    \vspace{-0.05in}
    \centering
    \resizebox{\textwidth}{!}{
    \renewcommand{\arraystretch}{0.95}
    \begin{tabular}{llccccc}
    \Xhline{0.2ex}
        \rule{0pt}{10pt}Method & Task & Validity (\%) & Uniqueness (\%) & \cellcolor{gray!25} Quality (\%) & Diversity & Distance \\
    \hline
        \rule{0pt}{10pt}DiGress & Linker design & \phantom{0}31.2~$\pm$~1.2 & 84.3~$\pm$~0.4 & \cellcolor{gray!25} \phantom{0}6.1~$\pm$~0.2 & 0.745~$\pm$~0.001 & 0.724~$\pm$~0.003 \\
        & Scaffold morphing & \phantom{0}31.2~$\pm$~1.2 & 84.3~$\pm$~0.4 & \cellcolor{gray!25} \phantom{0}6.1~$\pm$~0.2 & 0.745~$\pm$~0.001 & 0.724~$\pm$~0.003 \\
        & Motif extension & \phantom{0}21.8~$\pm$~0.8 & 94.5~$\pm$~0.3 & \cellcolor{gray!25} \phantom{0}4.2~$\pm$~0.1 & 0.818~$\pm$~0.003 & 0.794~$\pm$~0.003 \\
        & Scaffold decoration & \phantom{0}29.3~$\pm$~0.7 & 91.0~$\pm$~0.8 & \cellcolor{gray!25} \phantom{0}9.1~$\pm$~0.4 & 0.793~$\pm$~0.003 & 0.785~$\pm$~0.002 \\
        & Superstructure generation & \phantom{0}26.7~$\pm$~1.3 & 85.9~$\pm$~1.4 & \cellcolor{gray!25} \phantom{0}7.4~$\pm$~0.9 & 0.789~$\pm$~0.005 & 0.776~$\pm$~0.004 \\
    \hline
        \rule{0pt}{10pt}GenMol w/o MCG & Linker design & 100.0~$\pm$~0.0 & 83.7~$\pm$~0.5 & \cellcolor{gray!25} 21.9~$\pm$~0.4 & 0.547~$\pm$~0.002 & 0.563~$\pm$~0.003 \\
        & Scaffold morphing & 100.0~$\pm$~0.0 & 83.7~$\pm$~0.5 & \cellcolor{gray!25} 21.9~$\pm$~0.4 & 0.547~$\pm$~0.002 & 0.563~$\pm$~0.003 \\
        & Motif extension & \phantom{0}77.2~$\pm$~0.1 & 77.8~$\pm$~0.2 & \cellcolor{gray!25} 27.5~$\pm$~0.8 & 0.617~$\pm$~0.002 & 0.682~$\pm$~0.002 \\
        & Scaffold decoration & \phantom{0}96.8~$\pm$~0.2 & 78.0~$\pm$~1.2 & \cellcolor{gray!25} 29.6~$\pm$~0.8 & 0.576~$\pm$~0.001 & 0.650~$\pm$~0.001 \\
        & Superstructure generation & \phantom{0}98.2~$\pm$~1.1 & 78.3~$\pm$~3.4 & \cellcolor{gray!25} 33.3~$\pm$~1.6 & 0.574~$\pm$~0.008 & 0.757~$\pm$~0.003 \\
    \hline
        \rule{0pt}{10pt}GenMol & Linker design & 100.0~$\pm$~0.0 \textcolor{violet}{\scriptsize{(+0.0)}} & 83.7~$\pm$~0.5 \textcolor{violet}{\scriptsize{(+0.0)}} & \cellcolor{gray!25} 21.9~$\pm$~0.4 \textcolor{violet}{\scriptsize{(+0.0)}} & 0.547~$\pm$~0.002 \textcolor{violet}{\scriptsize{(+0.000)}} & 0.563~$\pm$~0.003 \textcolor{violet}{\scriptsize{(+0.000)}} \\
        & Scaffold morphing & 100.0~$\pm$~0.0 \textcolor{violet}{\scriptsize{(+0.0)}} & 83.7~$\pm$~0.5 \textcolor{violet}{\scriptsize{(+0.0)}} & \cellcolor{gray!25} 21.9~$\pm$~0.4 \textcolor{violet}{\scriptsize{(+0.0)}} & 0.547~$\pm$~0.002 \textcolor{violet}{\scriptsize{(+0.000)}} & 0.563~$\pm$~0.003 \textcolor{violet}{\scriptsize{(+0.000)}} \\
        & Motif extension & \phantom{0}82.9~$\pm$~0.1 \textcolor{violet}{\scriptsize{(+5.7)}} & 77.5~$\pm$~0.1 \textcolor{violet}{\scriptsize{(-0.3)}} & \cellcolor{gray!25} 30.1~$\pm$~0.4 \textcolor{violet}{\scriptsize{(+2.6)}} & 0.617~$\pm$~0.002 \textcolor{violet}{\scriptsize{(+0.000)}} & 0.682~$\pm$~0.001 \textcolor{violet}{\scriptsize{(+0.000)}} \\
        & Scaffold decoration & \phantom{0}96.6~$\pm$~0.8 \textcolor{violet}{\scriptsize{(-0.2)}} & 82.7~$\pm$~1.8 \textcolor{violet}{\scriptsize{(+4.7)}} & \cellcolor{gray!25} 31.8~$\pm$~0.5 \textcolor{violet}{\scriptsize{(+2.2)}} & 0.591~$\pm$~0.001 \textcolor{violet}{\scriptsize{(+0.015)}} & 0.651~$\pm$~0.001 \textcolor{violet}{\scriptsize{(+0.001)}} \\
        & Superstructure generation & \phantom{0}97.5~$\pm$~0.9 \textcolor{violet}{\scriptsize{(-0.7)}} & 83.6~$\pm$~1.0 \textcolor{violet}{\scriptsize{(+5.3)}} & \cellcolor{gray!25} 34.8~$\pm$~1.0 \textcolor{violet}{\scriptsize{(+1.5)}} & 0.599~$\pm$~0.009 \textcolor{violet}{\scriptsize{(+0.025)}} & 0.762~$\pm$~0.007 \textcolor{violet}{\scriptsize{(+0.005)}} \\
    \Xhline{0.2ex}
    \end{tabular}}
    \label{tab:app_frag}
\end{table*}

\subsection{Full Goal-directed Hit Generation Results}
We provide the full results of Table~\ref{tab:pmo} including the additional baselines from \citet{gao2022sample} in Table~\ref{tab:app_pmo}, Table~\ref{tab:app_pmo_2}, and Table~\ref{tab:app_pmo_3}. As shown in the tables, GenMol outperforms all baselines by a large margin.

On the other hand, to simulate the fragment-based drug discovery (FBDD) scenario following \citet{lee2024frag}, we have assumed that a high-quality fragment vocabulary of size 100 is given for each task in Table~\ref{tab:pmo}. Here, similar to \citet{wang2023retrieval}, we also provide the results of GenMol (10 fragments), which starts with only 10 of these fragments to simulate a scenario where high-quality fragments are sparse, in Table~\ref{tab:app_pmo}.

\subsection{Full Goal-directed Hit Generation Results in Ablation Study}
We provide the full results of the ablated GenMol variants baselines in the goal-directed hit generatino task in Table~\ref{tab:pmo_ablation} in Table~\ref{tab:app_pmo_ablation}. As shown in the table, A, A + Token remasking, A + GPT remasking and A + Fragment remasking show inferior performance to A + F + MCG, the full GenMol, as discussed in Section~\ref{sec:exp_ablation}.

\subsection{Ablation Study in Goal-directed Lead Optimization}
We compare the results of the ablated GenMol baselines in the goal-directed lead optimization task in Table~\ref{tab:app_lead_ablation}. As in Table~\ref{tab:pmo_ablation}, A + Token remasking, A + GPT remasking and A + Fragment remasking all outperform A, demonstrating the importance of further modification on top of A through remasking. Moreover, A + Fragment remasking generally outperforms A + Token remasking and A + GPT remasking, showing the effectiveness of the proposed remasking strategy that uses fragments as the exploration unit under discrete diffusion. Lastly, A + F + MCG shows further improved performance compared to A + F, validating the effectiveness of MCG in the lead optimization task.

\subsection{Examples of Generated Molecules}
\begin{figure}[t!]
    \centering
    \includegraphics[width=\linewidth]{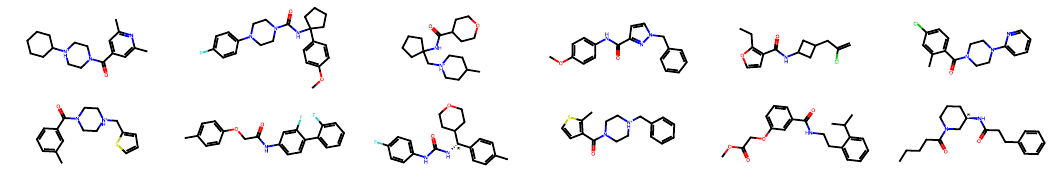}
    \caption{\small \textbf{Examples of generated molecules on \emph{de novo} generation.}}
    \label{fig:mols_denovo}
\end{figure}

\begin{figure}[t!]
    \centering
    \includegraphics[width=\linewidth]{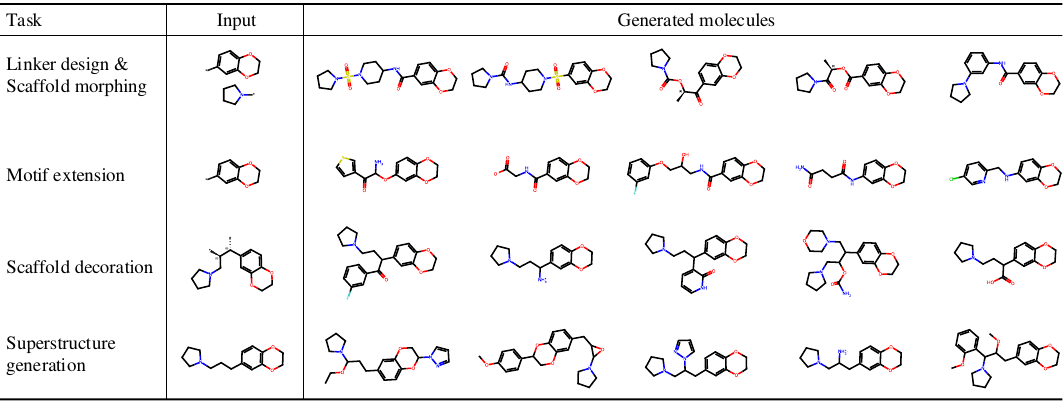}
    \caption{\small \textbf{Examples of generated molecules on fragment-constrained generation} of Eliglustat.}
    \label{fig:mols_frag}
\end{figure}

We provide examples of the molecules generated by GenMol ($N=1$, $\tau=0.5$, $r=0.5$) on \emph{de novo} generation Figure~\ref{fig:mols_denovo}. We provide examples of generated molecules on fragment-constrained generation in Figure~\ref{fig:mols_frag}.

\clearpage
\begin{table*}[t!]
    \centering
    \caption{\small \textbf{Goal-directed hit generation results.} The results are the means and standard deviations of PMO AUC top-10 of 3 runs. The results for $f$-RAG~\citep{lee2024frag}, Genetic GFN~\citep{kim2024genetic} and Mol GA~\citep{tripp2023genetic} are taken from the respective papers and the results for other baselines are taken from \citet{gao2022sample}. The best results are highlighted in bold.}
    \resizebox{0.9\textwidth}{!}{
    \begin{tabular}{l|ccccc}
    \Xhline{0.2ex}
        \rule{0pt}{15pt}Oracle & GenMol & \makecell{GenMol\\(10 fragments)} & $f$-RAG & Genetic GFN & Mol GA \\
    \hline
        \rule{0pt}{10pt}albuterol\_similarity & 0.937 $\pm$ 0.010 & 0.847 $\pm$ 0.036 & \textbf{0.977} $\pm$ 0.002 & 0.949 $\pm$ 0.010 & 0.896 $\pm$ 0.035 \\
        amlodipine\_mpo & \textbf{0.810} $\pm$ 0.012 & 0.762 $\pm$ 0.012 & 0.749 $\pm$ 0.019 & 0.761 $\pm$ 0.019 & 0.688 $\pm$ 0.039 \\
        celecoxib\_rediscovery & \textbf{0.826} $\pm$ 0.018 & 0.619 $\pm$ 0.005 & 0.778 $\pm$ 0.007 & 0.802 $\pm$ 0.029 & 0.567 $\pm$ 0.083 \\
        deco\_hop & \textbf{0.960} $\pm$ 0.010 & 0.957 $\pm$ 0.016 & 0.936 $\pm$ 0.011 & 0.733 $\pm$ 0.109 & 0.649 $\pm$ 0.025 \\
        drd2 & \textbf{0.995} $\pm$ 0.000 & \textbf{0.995} $\pm$ 0.000 & 0.992 $\pm$ 0.000 & 0.974 $\pm$ 0.006 & 0.936 $\pm$ 0.016 \\
        fexofenadine\_mpo & \textbf{0.894} $\pm$ 0.028 & 0.806 $\pm$ 0.008 & 0.856 $\pm$ 0.016 & 0.856 $\pm$ 0.039 & 0.825 $\pm$ 0.019 \\
        gsk3b & \textbf{0.986} $\pm$ 0.003 & 0.985 $\pm$ 0.004 & 0.969 $\pm$ 0.003 & 0.881 $\pm$ 0.042 & 0.843 $\pm$ 0.039 \\
        isomers\_c7h8n2o2 & 0.942 $\pm$ 0.004 & \textbf{0.984} $\pm$ 0.002 & 0.955 $\pm$ 0.008 & 0.969 $\pm$ 0.003 & 0.878 $\pm$ 0.026 \\
        isomers\_c9h10n2o2pf2cl & 0.833 $\pm$ 0.014 & 0.866 $\pm$ 0.010 & 0.850 $\pm$ 0.005 & \textbf{0.897} $\pm$ 0.007 & 0.865 $\pm$ 0.012 \\
        jnk3 & \textbf{0.906} $\pm$ 0.023 & 0.828 $\pm$ 0.007 & 0.904 $\pm$ 0.004 & 0.764 $\pm$ 0.069 & 0.702 $\pm$ 0.123 \\
        median1 & \textbf{0.398} $\pm$ 0.000 & 0.336 $\pm$ 0.008 & 0.340 $\pm$ 0.007 & 0.379 $\pm$ 0.010 & 0.257 $\pm$ 0.009 \\
        median2 & \textbf{0.359} $\pm$ 0.004 & 0.354 $\pm$ 0.000 & 0.323 $\pm$ 0.005 & 0.294 $\pm$ 0.007 & 0.301 $\pm$ 0.021 \\
        mestranol\_similarity & 0.982 $\pm$ 0.000 & \textbf{0.991} $\pm$ 0.002 & 0.671 $\pm$ 0.021 & 0.708 $\pm$ 0.057 & 0.591 $\pm$ 0.053 \\
        osimertinib\_mpo & \textbf{0.876} $\pm$ 0.008 & 0.870 $\pm$ 0.004 & 0.866 $\pm$ 0.009 & 0.860 $\pm$ 0.008 & 0.844 $\pm$ 0.015 \\
        perindopril\_mpo & \textbf{0.718} $\pm$ 0.012 & 0.695 $\pm$ 0.004 & 0.681 $\pm$ 0.017 & 0.595 $\pm$ 0.014 & 0.547 $\pm$ 0.022 \\
        qed & 0.942 $\pm$ 0.000 & \textbf{0.943} $\pm$ 0.000 & 0.939 $\pm$ 0.001 & \textbf{0.942} $\pm$ 0.000 & 0.941 $\pm$ 0.001 \\
        ranolazine\_mpo & \textbf{0.821} $\pm$ 0.011 & 0.777 $\pm$ 0.016 & 0.820 $\pm$ 0.016 & 0.819 $\pm$ 0.018 & 0.804 $\pm$ 0.011 \\
        scaffold\_hop & 0.628 $\pm$ 0.008 & \textbf{0.648} $\pm$ 0.005 & 0.576 $\pm$ 0.014 & 0.615 $\pm$ 0.100 & 0.527 $\pm$ 0.025 \\
        sitagliptin\_mpo & 0.584 $\pm$ 0.034 & 0.588 $\pm$ 0.064 & 0.601 $\pm$ 0.011 & \textbf{0.634} $\pm$ 0.039 & 0.582 $\pm$ 0.040 \\
        thiothixene\_rediscovery & \textbf{0.692} $\pm$ 0.123 & 0.569 $\pm$ 0.013 & 0.584 $\pm$ 0.009 & 0.583 $\pm$ 0.034 & 0.519 $\pm$ 0.041 \\
        troglitazone\_rediscovery & \textbf{0.867} $\pm$ 0.022 & 0.848 $\pm$ 0.040 & 0.448 $\pm$ 0.017 & 0.511 $\pm$ 0.054 & 0.427 $\pm$ 0.031 \\
        valsartan\_smarts & \textbf{0.822} $\pm$ 0.042 & 0.803 $\pm$ 0.011 & 0.627 $\pm$ 0.058 & 0.135 $\pm$ 0.271 & 0.000 $\pm$ 0.000 \\
        zaleplon\_mpo & \textbf{0.584} $\pm$ 0.011 & 0.571 $\pm$ 0.016 & 0.486 $\pm$ 0.004 & 0.552 $\pm$ 0.033 & 0.519 $\pm$ 0.029 \\
    \hline
        \rule{0pt}{10pt}Sum & \textbf{18.362} & 17.643 & 16.928 & 16.213 & 14.708 \\
    \Xhline{0.2ex}
    \end{tabular}}
    
    \vspace{0.1in}
    \resizebox{0.9\textwidth}{!}{
    \begin{tabular}{l|ccccc}
    \Xhline{0.2ex}
        \rule{0pt}{15pt}Oracle & REINVENT & Graph GA & \makecell{SELFIES-\\REINVENT} & GP BO & STONED \\
    \hline
        \rule{0pt}{10pt}albuterol\_similarity & 0.882 $\pm$ 0.006 & 0.838 $\pm$ 0.016 & 0.826 $\pm$ 0.030 & 0.898 $\pm$ 0.014 & 0.745 $\pm$ 0.076 \\
        amlodipine\_mpo & 0.635 $\pm$ 0.035 & 0.661 $\pm$ 0.020 & 0.607 $\pm$ 0.014 & 0.583 $\pm$ 0.044 & 0.608 $\pm$ 0.046 \\
        celecoxib\_rediscovery & 0.713 $\pm$ 0.067 & 0.630 $\pm$ 0.097 & 0.573 $\pm$ 0.043 & 0.723 $\pm$ 0.053 & 0.382 $\pm$ 0.041 \\
        deco\_hop & 0.666 $\pm$ 0.044 & 0.619 $\pm$ 0.004 & 0.631 $\pm$ 0.012 & 0.629 $\pm$ 0.018 & 0.611 $\pm$ 0.008 \\
        drd2 & 0.945 $\pm$ 0.007 & 0.964 $\pm$ 0.012 & 0.943 $\pm$ 0.005 & 0.923 $\pm$ 0.017 & 0.913 $\pm$ 0.020 \\
        fexofenadine\_mpo & 0.784 $\pm$ 0.006 & 0.760 $\pm$ 0.011 & 0.741 $\pm$ 0.002 & 0.722 $\pm$ 0.005 & 0.797 $\pm$ 0.016 \\
        gsk3b & 0.865 $\pm$ 0.043 & 0.788 $\pm$ 0.070 & 0.780 $\pm$ 0.037 & 0.851 $\pm$ 0.041 & 0.668 $\pm$ 0.049 \\
        isomers\_c7h8n2o2 & 0.852 $\pm$ 0.036 & 0.862 $\pm$ 0.065 & 0.849 $\pm$ 0.034 & 0.680 $\pm$ 0.117 & 0.899 $\pm$ 0.011 \\
        isomers\_c9h10n2o2pf2cl & 0.642 $\pm$ 0.054 & 0.719 $\pm$ 0.047 & 0.733 $\pm$ 0.029 & 0.469 $\pm$ 0.180 & 0.805 $\pm$ 0.031 \\
        jnk3 & 0.783 $\pm$ 0.023 & 0.553 $\pm$ 0.136 & 0.631 $\pm$ 0.064 & 0.564 $\pm$ 0.155 & 0.523 $\pm$ 0.092 \\
        median1 & 0.356 $\pm$ 0.009 & 0.294 $\pm$ 0.021 & 0.355 $\pm$ 0.011 & 0.301 $\pm$ 0.014 & 0.266 $\pm$ 0.016 \\
        median2 & 0.276 $\pm$ 0.008 & 0.273 $\pm$ 0.009 & 0.255 $\pm$ 0.005 & 0.297 $\pm$ 0.009 & 0.245 $\pm$ 0.032 \\
        mestranol\_similarity & 0.618 $\pm$ 0.048 & 0.579 $\pm$ 0.022 & 0.620 $\pm$ 0.029 & 0.627 $\pm$ 0.089 & 0.609 $\pm$ 0.101 \\
        osimertinib\_mpo & 0.837 $\pm$ 0.009 & 0.831 $\pm$ 0.005 & 0.820 $\pm$ 0.003 & 0.787 $\pm$ 0.006 & 0.822 $\pm$ 0.012 \\
        perindopril\_mpo & 0.537 $\pm$ 0.016 & 0.538 $\pm$ 0.009 & 0.517 $\pm$ 0.021 & 0.493 $\pm$ 0.011 & 0.488 $\pm$ 0.011 \\
        qed & 0.941 $\pm$ 0.000 & 0.940 $\pm$ 0.000 & 0.940 $\pm$ 0.000 & 0.937 $\pm$ 0.000 & 0.941 $\pm$ 0.000 \\
        ranolazine\_mpo & 0.760 $\pm$ 0.009 & 0.728 $\pm$ 0.012 & 0.748 $\pm$ 0.018 & 0.735 $\pm$ 0.013 & 0.765 $\pm$ 0.029 \\
        scaffold\_hop & 0.560 $\pm$ 0.019 & 0.517 $\pm$ 0.007 & 0.525 $\pm$ 0.013 & 0.548 $\pm$ 0.019 & 0.521 $\pm$ 0.034 \\
        sitagliptin\_mpo & 0.021 $\pm$ 0.003 & 0.433 $\pm$ 0.075 & 0.194 $\pm$ 0.121 & 0.186 $\pm$ 0.055 & 0.393 $\pm$ 0.083 \\
        thiothixene\_rediscovery & 0.534 $\pm$ 0.013 & 0.479 $\pm$ 0.025 & 0.495 $\pm$ 0.040 & 0.559 $\pm$ 0.027 & 0.367 $\pm$ 0.027 \\
        troglitazone\_rediscovery & 0.441 $\pm$ 0.032 & 0.390 $\pm$ 0.016 & 0.348 $\pm$ 0.012 & 0.410 $\pm$ 0.015 & 0.320 $\pm$ 0.018 \\
        valsartan\_smarts & 0.179 $\pm$ 0.358 & 0.000 $\pm$ 0.000 & 0.000 $\pm$ 0.000 & 0.000 $\pm$ 0.000 & 0.000 $\pm$ 0.000 \\
        zaleplon\_mpo & 0.358 $\pm$ 0.062 & 0.346 $\pm$ 0.032 & 0.333 $\pm$ 0.026 & 0.221 $\pm$ 0.072 & 0.325 $\pm$ 0.027 \\
    \hline
        \rule{0pt}{10pt}Sum & 14.196 & 13.751 & 13.471 & 13.156 & 13.024 \\
    \Xhline{0.2ex}
    \end{tabular}}
    \label{tab:app_pmo}
\end{table*}

\begin{table*}[t!]
    \centering
    \caption{\small \textbf{Goal-directed hit generation results} (continued).}
    \resizebox{0.9\textwidth}{!}{
    \begin{tabular}{l|ccccc}
    \Xhline{0.2ex}
        \rule{0pt}{10pt}Oracle & LSTM HC & SMILES-GA & SynNet & DoG-Gen & DST \\
    \hline
        \rule{0pt}{10pt}albuterol\_similarity & 0.719 $\pm$ 0.018 & 0.661 $\pm$ 0.066 & 0.584 $\pm$ 0.039 & 0.676 $\pm$ 0.013 & 0.619 $\pm$ 0.020 \\
        amlodipine\_mpo & 0.593 $\pm$ 0.016 & 0.549 $\pm$ 0.009 & 0.565 $\pm$ 0.007 & 0.536 $\pm$ 0.003 & 0.516 $\pm$ 0.007 \\
        celecoxib\_rediscovery & 0.539 $\pm$ 0.018 & 0.344 $\pm$ 0.027 & 0.441 $\pm$ 0.027 & 0.464 $\pm$ 0.009 & 0.380 $\pm$ 0.006 \\
        deco\_hop & 0.826 $\pm$ 0.017 & 0.611 $\pm$ 0.006 & 0.613 $\pm$ 0.009 & 0.800 $\pm$ 0.007 & 0.608 $\pm$ 0.008 \\
        drd2 & 0.919 $\pm$ 0.015 & 0.908 $\pm$ 0.019 & 0.969 $\pm$ 0.004 & 0.948 $\pm$ 0.001 & 0.820 $\pm$ 0.014 \\
        fexofenadine\_mpo & 0.725 $\pm$ 0.003 & 0.721 $\pm$ 0.015 & 0.761 $\pm$ 0.015 & 0.695 $\pm$ 0.003 & 0.725 $\pm$ 0.005 \\
        gsk3b & 0.839 $\pm$ 0.015 & 0.629 $\pm$ 0.044 & 0.789 $\pm$ 0.032 & 0.831 $\pm$ 0.021 & 0.671 $\pm$ 0.032 \\
        isomers\_c7h8n2o2 & 0.485 $\pm$ 0.045 & 0.913 $\pm$ 0.021 & 0.455 $\pm$ 0.031 & 0.465 $\pm$ 0.018 & 0.548 $\pm$ 0.069 \\
        isomers\_c9h10n2o2pf2cl & 0.342 $\pm$ 0.027 & 0.860 $\pm$ 0.065 & 0.241 $\pm$ 0.064 & 0.199 $\pm$ 0.016 & 0.458 $\pm$ 0.063 \\
        jnk3 & 0.661 $\pm$ 0.039 & 0.316 $\pm$ 0.022 & 0.630 $\pm$ 0.034 & 0.595 $\pm$ 0.023 & 0.556 $\pm$ 0.057 \\
        median1 & 0.255 $\pm$ 0.010 & 0.192 $\pm$ 0.012 & 0.218 $\pm$ 0.008 & 0.217 $\pm$ 0.001 & 0.232 $\pm$ 0.009 \\
        median2 & 0.248 $\pm$ 0.008 & 0.198 $\pm$ 0.005 & 0.235 $\pm$ 0.006 & 0.212 $\pm$ 0.000 & 0.185 $\pm$ 0.020 \\
        mestranol\_similarity & 0.526 $\pm$ 0.032 & 0.469 $\pm$ 0.029 & 0.399 $\pm$ 0.021 & 0.437 $\pm$ 0.007 & 0.450 $\pm$ 0.027 \\
        osimertinib\_mpo & 0.796 $\pm$ 0.002 & 0.817 $\pm$ 0.011 & 0.796 $\pm$ 0.003 & 0.774 $\pm$ 0.002 & 0.785 $\pm$ 0.004 \\
        perindopril\_mpo & 0.489 $\pm$ 0.007 & 0.447 $\pm$ 0.013 & 0.557 $\pm$ 0.011 & 0.474 $\pm$ 0.002 & 0.462 $\pm$ 0.008 \\
        qed & 0.939 $\pm$ 0.000 & 0.940 $\pm$ 0.000 & 0.941 $\pm$ 0.000 & 0.934 $\pm$ 0.000 & 0.938 $\pm$ 0.000 \\
        ranolazine\_mpo & 0.714 $\pm$ 0.008 & 0.699 $\pm$ 0.026 & 0.741 $\pm$ 0.010 & 0.711 $\pm$ 0.006 & 0.632 $\pm$ 0.054 \\
        scaffold\_hop & 0.533 $\pm$ 0.012 & 0.494 $\pm$ 0.011 & 0.502 $\pm$ 0.012 & 0.515 $\pm$ 0.005 & 0.497 $\pm$ 0.004 \\
        sitagliptin\_mpo & 0.066 $\pm$ 0.019 & 0.363 $\pm$ 0.057 & 0.025 $\pm$ 0.014 & 0.048 $\pm$ 0.008 & 0.075 $\pm$ 0.032 \\
        thiothixene\_rediscovery & 0.438 $\pm$ 0.008 & 0.315 $\pm$ 0.017 & 0.401 $\pm$ 0.019 & 0.375 $\pm$ 0.004 & 0.366 $\pm$ 0.006 \\
        troglitazone\_rediscovery & 0.354 $\pm$ 0.016 & 0.263 $\pm$ 0.024 & 0.283 $\pm$ 0.008 & 0.416 $\pm$ 0.019 & 0.279 $\pm$ 0.019 \\
        valsartan\_smarts & 0.000 $\pm$ 0.000 & 0.000 $\pm$ 0.000 & 0.000 $\pm$ 0.000 & 0.000 $\pm$ 0.000 & 0.000 $\pm$ 0.000 \\
        zaleplon\_mpo & 0.206 $\pm$ 0.006 & 0.334 $\pm$ 0.041 & 0.341 $\pm$ 0.011 & 0.123 $\pm$ 0.016 & 0.176 $\pm$ 0.045 \\
    \hline
        \rule{0pt}{10pt}Sum & 12.223 & 12.054 & 11.498 & 11.456 & 10.989 \\
    \Xhline{0.2ex}
    \end{tabular}}
    
    \vspace{0.1in}
    \resizebox{0.9\textwidth}{!}{
    \begin{tabular}{l|ccccc}
    \Xhline{0.2ex}
        \rule{0pt}{15pt}Oracle & MARS & MIMOSA & MolPal & \makecell{SELFIES-\\LSTM HC} & DoG-AE \\
    \hline
        \rule{0pt}{10pt}albuterol\_similarity & 0.597 $\pm$ 0.124 & 0.618 $\pm$ 0.017 & 0.609 $\pm$ 0.002 & 0.664 $\pm$ 0.030 & 0.533 $\pm$ 0.034 \\
        amlodipine\_mpo & 0.504 $\pm$ 0.016 & 0.543 $\pm$ 0.003 & 0.582 $\pm$ 0.008 & 0.532 $\pm$ 0.004 & 0.507 $\pm$ 0.005 \\
        celecoxib\_rediscovery & 0.379 $\pm$ 0.060 & 0.393 $\pm$ 0.010 & 0.415 $\pm$ 0.001 & 0.385 $\pm$ 0.008 & 0.355 $\pm$ 0.012 \\
        deco\_hop & 0.589 $\pm$ 0.003 & 0.619 $\pm$ 0.003 & 0.643 $\pm$ 0.005 & 0.590 $\pm$ 0.001 & 0.765 $\pm$ 0.055 \\
        drd2 & 0.891 $\pm$ 0.020 & 0.799 $\pm$ 0.017 & 0.783 $\pm$ 0.009 & 0.729 $\pm$ 0.034 & 0.943 $\pm$ 0.009 \\
        fexofenadine\_mpo & 0.711 $\pm$ 0.006 & 0.706 $\pm$ 0.011 & 0.685 $\pm$ 0.000 & 0.693 $\pm$ 0.004 & 0.679 $\pm$ 0.017 \\
        gsk3b & 0.552 $\pm$ 0.037 & 0.554 $\pm$ 0.042 & 0.555 $\pm$ 0.011 & 0.423 $\pm$ 0.018 & 0.601 $\pm$ 0.091 \\
        isomers\_c7h8n2o2 & 0.728 $\pm$ 0.027 & 0.564 $\pm$ 0.046 & 0.484 $\pm$ 0.006 & 0.587 $\pm$ 0.031 & 0.239 $\pm$ 0.077 \\
        isomers\_c9h10n2o2pf2cl & 0.581 $\pm$ 0.013 & 0.303 $\pm$ 0.046 & 0.164 $\pm$ 0.003 & 0.352 $\pm$ 0.019 & 0.049 $\pm$ 0.015 \\
        jnk3 & 0.489 $\pm$ 0.095 & 0.360 $\pm$ 0.063 & 0.339 $\pm$ 0.009 & 0.207 $\pm$ 0.013 & 0.469 $\pm$ 0.138 \\
        median1 & 0.207 $\pm$ 0.011 & 0.243 $\pm$ 0.005 & 0.249 $\pm$ 0.001 & 0.239 $\pm$ 0.009 & 0.171 $\pm$ 0.009 \\
        median2 & 0.181 $\pm$ 0.011 & 0.214 $\pm$ 0.002 & 0.230 $\pm$ 0.000 & 0.205 $\pm$ 0.005 & 0.182 $\pm$ 0.006 \\
        mestranol\_similarity & 0.388 $\pm$ 0.026 & 0.438 $\pm$ 0.015 & 0.564 $\pm$ 0.004 & 0.446 $\pm$ 0.009 & 0.370 $\pm$ 0.014 \\
        osimertinib\_mpo & 0.777 $\pm$ 0.006 & 0.788 $\pm$ 0.014 & 0.779 $\pm$ 0.000 & 0.780 $\pm$ 0.005 & 0.750 $\pm$ 0.012 \\
        perindopril\_mpo & 0.462 $\pm$ 0.006 & 0.490 $\pm$ 0.011 & 0.467 $\pm$ 0.002 & 0.448 $\pm$ 0.006 & 0.432 $\pm$ 0.013 \\
        qed & 0.930 $\pm$ 0.003 & 0.939 $\pm$ 0.000 & 0.940 $\pm$ 0.000 & 0.938 $\pm$ 0.000 & 0.926 $\pm$ 0.003 \\
        ranolazine\_mpo & 0.740 $\pm$ 0.010 & 0.640 $\pm$ 0.015 & 0.457 $\pm$ 0.005 & 0.614 $\pm$ 0.010 & 0.689 $\pm$ 0.015 \\
        scaffold\_hop & 0.469 $\pm$ 0.004 & 0.507 $\pm$ 0.015 & 0.494 $\pm$ 0.000 & 0.472 $\pm$ 0.002 & 0.489 $\pm$ 0.010 \\
        sitagliptin\_mpo & 0.016 $\pm$ 0.003 & 0.102 $\pm$ 0.023 & 0.043 $\pm$ 0.001 & 0.116 $\pm$ 0.012 & 0.009 $\pm$ 0.005 \\
        thiothixene\_rediscovery & 0.344 $\pm$ 0.022 & 0.347 $\pm$ 0.018 & 0.339 $\pm$ 0.001 & 0.339 $\pm$ 0.009 & 0.314 $\pm$ 0.015 \\
        troglitazone\_rediscovery & 0.256 $\pm$ 0.016 & 0.299 $\pm$ 0.009 & 0.268 $\pm$ 0.000 & 0.257 $\pm$ 0.002 & 0.259 $\pm$ 0.016 \\
        valsartan\_smarts & 0.000 $\pm$ 0.000 & 0.000 $\pm$ 0.000 & 0.000 $\pm$ 0.000 & 0.000 $\pm$ 0.000 & 0.000 $\pm$ 0.000 \\
        zaleplon\_mpo & 0.187 $\pm$ 0.046 & 0.172 $\pm$ 0.036 & 0.168 $\pm$ 0.003 & 0.218 $\pm$ 0.020 & 0.049 $\pm$ 0.027 \\
    \hline
        \rule{0pt}{10pt}Sum & 10.989 & 10.651 & 10.268 & 10.246 & 9.790 \\
    \Xhline{0.2ex}
    \end{tabular}}
    \label{tab:app_pmo_2}
\end{table*}

\begin{table*}[t!]
    \centering
    \caption{\small \textbf{Goal-directed hit generation results} (continued).}
    \resizebox{0.9\textwidth}{!}{
    \begin{tabular}{l|ccccc}
    \Xhline{0.2ex}
        \rule{0pt}{15pt}Oracle & GFlowNet & GA+D & \makecell{SELFIES-\\VAE BO} & Screening & \makecell{SMILES-\\VAE BO} \\
    \hline
        \rule{0pt}{10pt}albuterol\_similarity & 0.447 $\pm$ 0.012 & 0.495 $\pm$ 0.025 & 0.494 $\pm$ 0.012 & 0.483 $\pm$ 0.006 & 0.489 $\pm$ 0.007 \\
        amlodipine\_mpo & 0.444 $\pm$ 0.004 & 0.400 $\pm$ 0.032 & 0.516 $\pm$ 0.005 & 0.535 $\pm$ 0.001 & 0.533 $\pm$ 0.009 \\
        celecoxib\_rediscovery & 0.327 $\pm$ 0.004 & 0.223 $\pm$ 0.025 & 0.326 $\pm$ 0.007 & 0.351 $\pm$ 0.005 & 0.354 $\pm$ 0.002 \\
        deco\_hop & 0.583 $\pm$ 0.002 & 0.550 $\pm$ 0.005 & 0.579 $\pm$ 0.001 & 0.590 $\pm$ 0.001 & 0.589 $\pm$ 0.001 \\
        drd2 & 0.590 $\pm$ 0.070 & 0.382 $\pm$ 0.205 & 0.569 $\pm$ 0.039 & 0.545 $\pm$ 0.015 & 0.555 $\pm$ 0.043 \\
        fexofenadine\_mpo & 0.693 $\pm$ 0.006 & 0.587 $\pm$ 0.007 & 0.670 $\pm$ 0.004 & 0.666 $\pm$ 0.004 & 0.671 $\pm$ 0.003 \\
        gsk3b & 0.651 $\pm$ 0.026 & 0.342 $\pm$ 0.019 & 0.350 $\pm$ 0.034 & 0.438 $\pm$ 0.034 & 0.386 $\pm$ 0.006 \\
        isomers\_c7h8n2o2 & 0.366 $\pm$ 0.043 & 0.854 $\pm$ 0.015 & 0.325 $\pm$ 0.028 & 0.168 $\pm$ 0.034 & 0.161 $\pm$ 0.017 \\
        isomers\_c9h10n2o2pf2cl & 0.110 $\pm$ 0.031 & 0.657 $\pm$ 0.020 & 0.200 $\pm$ 0.030 & 0.106 $\pm$ 0.021 & 0.084 $\pm$ 0.009 \\
        jnk3 & 0.440 $\pm$ 0.022 & 0.219 $\pm$ 0.021 & 0.208 $\pm$ 0.022 & 0.238 $\pm$ 0.024 & 0.241 $\pm$ 0.026 \\
        median1 & 0.202 $\pm$ 0.004 & 0.180 $\pm$ 0.009 & 0.201 $\pm$ 0.003 & 0.205 $\pm$ 0.005 & 0.202 $\pm$ 0.006 \\
        median2 & 0.180 $\pm$ 0.000 & 0.121 $\pm$ 0.005 & 0.185 $\pm$ 0.001 & 0.200 $\pm$ 0.004 & 0.195 $\pm$ 0.001 \\
        mestranol\_similarity & 0.322 $\pm$ 0.007 & 0.371 $\pm$ 0.016 & 0.386 $\pm$ 0.009 & 0.409 $\pm$ 0.019 & 0.399 $\pm$ 0.005 \\
        osimertinib\_mpo & 0.784 $\pm$ 0.001 & 0.672 $\pm$ 0.027 & 0.765 $\pm$ 0.002 & 0.764 $\pm$ 0.001 & 0.771 $\pm$ 0.002 \\
        perindopril\_mpo & 0.430 $\pm$ 0.010 & 0.172 $\pm$ 0.088 & 0.429 $\pm$ 0.003 & 0.445 $\pm$ 0.004 & 0.442 $\pm$ 0.004 \\
        qed & 0.921 $\pm$ 0.004 & 0.860 $\pm$ 0.014 & 0.936 $\pm$ 0.001 & 0.938 $\pm$ 0.000 & 0.938 $\pm$ 0.000 \\
        ranolazine\_mpo & 0.652 $\pm$ 0.002 & 0.555 $\pm$ 0.015 & 0.452 $\pm$ 0.025 & 0.411 $\pm$ 0.010 & 0.457 $\pm$ 0.012 \\
        scaffold\_hop & 0.463 $\pm$ 0.002 & 0.413 $\pm$ 0.009 & 0.455 $\pm$ 0.004 & 0.471 $\pm$ 0.002 & 0.470 $\pm$ 0.003 \\
        sitagliptin\_mpo & 0.008 $\pm$ 0.003 & 0.281 $\pm$ 0.022 & 0.084 $\pm$ 0.015 & 0.022 $\pm$ 0.003 & 0.023 $\pm$ 0.004 \\
        thiothixene\_rediscovery & 0.285 $\pm$ 0.012 & 0.223 $\pm$ 0.029 & 0.297 $\pm$ 0.004 & 0.317 $\pm$ 0.003 & 0.317 $\pm$ 0.007 \\
        troglitazone\_rediscovery & 0.188 $\pm$ 0.001 & 0.152 $\pm$ 0.013 & 0.243 $\pm$ 0.004 & 0.249 $\pm$ 0.003 & 0.257 $\pm$ 0.003 \\
        valsartan\_smarts & 0.000 $\pm$ 0.000 & 0.000 $\pm$ 0.000 & 0.002 $\pm$ 0.003 & 0.000 $\pm$ 0.000 & 0.002 $\pm$ 0.004 \\
        zaleplon\_mpo & 0.035 $\pm$ 0.030 & 0.244 $\pm$ 0.015 & 0.206 $\pm$ 0.015 & 0.072 $\pm$ 0.014 & 0.039 $\pm$ 0.012 \\
    \hline
        \rule{0pt}{10pt}Sum & 9.131 & 8.964 & 8.887 & 8.635 & 8.587 \\
    \Xhline{0.2ex}
    \end{tabular}}
    
    \vspace{0.1in}
    \resizebox{0.9\textwidth}{!}{
    \begin{tabular}{l|ccccc}
    \Xhline{0.2ex}
        \rule{0pt}{10pt}Oracle & Pasithea & GFlowNet-AL & JT-VAE BO & Graph MCTS & MolDQN \\
    \hline
        \rule{0pt}{10pt}albuterol\_similarity & 0.447 $\pm$ 0.007 & 0.390 $\pm$ 0.008 & 0.485 $\pm$ 0.029 & 0.580 $\pm$ 0.023 & 0.320 $\pm$ 0.015 \\
        amlodipine\_mpo & 0.504 $\pm$ 0.003 & 0.428 $\pm$ 0.002 & 0.519 $\pm$ 0.009 & 0.447 $\pm$ 0.008 & 0.311 $\pm$ 0.008 \\
        celecoxib\_rediscovery & 0.312 $\pm$ 0.007 & 0.257 $\pm$ 0.003 & 0.299 $\pm$ 0.009 & 0.264 $\pm$ 0.013 & 0.099 $\pm$ 0.005 \\
        deco\_hop & 0.579 $\pm$ 0.001 & 0.583 $\pm$ 0.001 & 0.585 $\pm$ 0.002 & 0.554 $\pm$ 0.002 & 0.546 $\pm$ 0.001 \\
        drd2 & 0.255 $\pm$ 0.040 & 0.468 $\pm$ 0.046 & 0.506 $\pm$ 0.136 & 0.300 $\pm$ 0.050 & 0.025 $\pm$ 0.001 \\
        fexofenadine\_mpo & 0.660 $\pm$ 0.015 & 0.688 $\pm$ 0.002 & 0.667 $\pm$ 0.010 & 0.574 $\pm$ 0.009 & 0.478 $\pm$ 0.012 \\
        gsk3b & 0.281 $\pm$ 0.038 & 0.588 $\pm$ 0.015 & 0.350 $\pm$ 0.051 & 0.281 $\pm$ 0.022 & 0.241 $\pm$ 0.008 \\
        isomers\_c7h8n2o2 & 0.673 $\pm$ 0.030 & 0.241 $\pm$ 0.055 & 0.103 $\pm$ 0.016 & 0.530 $\pm$ 0.035 & 0.431 $\pm$ 0.035 \\
        isomers\_c9h10n2o2pf2cl & 0.345 $\pm$ 0.145 & 0.064 $\pm$ 0.012 & 0.090 $\pm$ 0.035 & 0.454 $\pm$ 0.067 & 0.342 $\pm$ 0.026 \\
        jnk3 & 0.154 $\pm$ 0.018 & 0.362 $\pm$ 0.021 & 0.222 $\pm$ 0.009 & 0.110 $\pm$ 0.019 & 0.111 $\pm$ 0.008 \\
        median1 & 0.178 $\pm$ 0.009 & 0.190 $\pm$ 0.002 & 0.179 $\pm$ 0.003 & 0.195 $\pm$ 0.005 & 0.122 $\pm$ 0.007 \\
        median2 & 0.179 $\pm$ 0.004 & 0.173 $\pm$ 0.001 & 0.180 $\pm$ 0.003 & 0.132 $\pm$ 0.002 & 0.088 $\pm$ 0.003 \\
        mestranol\_similarity & 0.361 $\pm$ 0.016 & 0.295 $\pm$ 0.004 & 0.356 $\pm$ 0.013 & 0.281 $\pm$ 0.008 & 0.188 $\pm$ 0.007 \\
        osimertinib\_mpo & 0.749 $\pm$ 0.007 & 0.787 $\pm$ 0.003 & 0.775 $\pm$ 0.004 & 0.700 $\pm$ 0.004 & 0.674 $\pm$ 0.006 \\
        perindopril\_mpo & 0.421 $\pm$ 0.008 & 0.421 $\pm$ 0.002 & 0.430 $\pm$ 0.009 & 0.277 $\pm$ 0.013 & 0.213 $\pm$ 0.043 \\
        qed & 0.931 $\pm$ 0.002 & 0.902 $\pm$ 0.005 & 0.934 $\pm$ 0.002 & 0.892 $\pm$ 0.006 & 0.731 $\pm$ 0.018 \\
        ranolazine\_mpo & 0.347 $\pm$ 0.012 & 0.632 $\pm$ 0.007 & 0.508 $\pm$ 0.055 & 0.239 $\pm$ 0.027 & 0.051 $\pm$ 0.020 \\
        scaffold\_hop & 0.456 $\pm$ 0.003 & 0.460 $\pm$ 0.002 & 0.470 $\pm$ 0.005 & 0.412 $\pm$ 0.003 & 0.405 $\pm$ 0.004 \\
        sitagliptin\_mpo & 0.088 $\pm$ 0.013 & 0.006 $\pm$ 0.001 & 0.046 $\pm$ 0.027 & 0.056 $\pm$ 0.012 & 0.003 $\pm$ 0.002 \\
        thiothixene\_rediscovery & 0.288 $\pm$ 0.006 & 0.266 $\pm$ 0.005 & 0.282 $\pm$ 0.008 & 0.231 $\pm$ 0.004 & 0.099 $\pm$ 0.007 \\
        troglitazone\_rediscovery & 0.240 $\pm$ 0.002 & 0.186 $\pm$ 0.003 & 0.237 $\pm$ 0.005 & 0.224 $\pm$ 0.009 & 0.122 $\pm$ 0.004 \\
        valsartan\_smarts & 0.006 $\pm$ 0.012 & 0.000 $\pm$ 0.000 & 0.000 $\pm$ 0.000 & 0.000 $\pm$ 0.000 & 0.000 $\pm$ 0.000 \\
        zaleplon\_mpo & 0.091 $\pm$ 0.013 & 0.010 $\pm$ 0.001 & 0.125 $\pm$ 0.038 & 0.058 $\pm$ 0.019 & 0.010 $\pm$ 0.005 \\
    \hline
        \rule{0pt}{10pt}Sum & 8.556 & 8.406 & 8.358 & 7.803 & 5.620 \\
    \Xhline{0.2ex}
    \end{tabular}}
    \label{tab:app_pmo_3}
\end{table*}

\clearpage
\begin{table*}[t!]
    \centering
    \caption{\small \textbf{Ablation study on goal-directed hit generation.} The results are the means and standard deviations of PMO AUC top-10 of 3 runs. The best results are highlighted in bold.}
    \resizebox{\textwidth}{!}{
    \renewcommand{\tabcolsep}{3mm}
    \begin{tabular}{l|ccccc}
    \Xhline{0.2ex}
        \rule{0pt}{10pt}Oracle & Attaching (A) & \makecell{A +\\Token remasking} & \makecell{A +\\GPT remasking} & \makecell{A +\\Frag. remasking (F)} & A + F + MCG \\
    \hline
        \rule{0pt}{10pt}albuterol\_similarity & 0.872 $\pm$ 0.032 & 0.895 $\pm$ 0.033 & 0.908 $\pm$ 0.039 & 0.932 $\pm$ 0.007 & \textbf{0.937} $\pm$ 0.010 \\
        amlodipine\_mpo & 0.769 $\pm$ 0.029 & 0.802 $\pm$ 0.016 & 0.780 $\pm$ 0.032 & 0.804 $\pm$ 0.006 & \textbf{0.810} $\pm$ 0.012 \\
        celecoxib\_rediscovery & \textbf{0.859} $\pm$ 0.008 & 0.821 $\pm$ 0.010 & 0.847 $\pm$ 0.006 & 0.826 $\pm$ 0.018 & 0.826 $\pm$ 0.018 \\
        deco\_hop & 0.917 $\pm$ 0.009 & 0.945 $\pm$ 0.006 & 0.955 $\pm$ 0.005 & 0.953 $\pm$ 0.016 & \textbf{0.960} $\pm$ 0.010 \\
        drd2 & \textbf{0.995} $\pm$ 0.000 & \textbf{0.995} $\pm$ 0.000 & \textbf{0.995} $\pm$ 0.000 & \textbf{0.995} $\pm$ 0.000 & \textbf{0.995} $\pm$ 0.000 \\
        fexofenadine\_mpo & 0.875 $\pm$ 0.019 & 0.886 $\pm$ 0.017 & \textbf{0.905} $\pm$ 0.012 & 0.894 $\pm$ 0.028 & 0.894 $\pm$ 0.028 \\
        gsk3b & 0.985 $\pm$ 0.003 & 0.985 $\pm$ 0.003 & \textbf{0.986} $\pm$ 0.003 & \textbf{0.986} $\pm$ 0.003 & \textbf{0.986} $\pm$ 0.001 \\
        isomers\_c7h8n2o2 & 0.897 $\pm$ 0.016 & 0.934 $\pm$ 0.003 & 0.915 $\pm$ 0.008 & 0.934 $\pm$ 0.002 & \textbf{0.942} $\pm$ 0.004 \\
        isomers\_c9h10n2o2pf2cl & 0.816 $\pm$ 0.025 & 0.830 $\pm$ 0.016 & 0.820 $\pm$ 0.018 & \textbf{0.833} $\pm$ 0.014 & \textbf{0.833} $\pm$ 0.014 \\
        jnk3 & 0.845 $\pm$ 0.035 & 0.848 $\pm$ 0.016 & 0.840 $\pm$ 0.021 & 0.856 $\pm$ 0.016 & \textbf{0.906} $\pm$ 0.023 \\
        median1 & 0.397 $\pm$ 0.000 & 0.397 $\pm$ 0.000 & 0.396 $\pm$ 0.000 & 0.397 $\pm$ 0.000 & \textbf{0.398} $\pm$ 0.000 \\
        median2 & 0.349 $\pm$ 0.004 & 0.350 $\pm$ 0.006 & 0.353 $\pm$ 0.004 & 0.355 $\pm$ 0.003 & \textbf{0.359} $\pm$ 0.004 \\
        mestranol\_similarity & 0.970 $\pm$ 0.004 & 0.980 $\pm$ 0.002 & 0.980 $\pm$ 0.002 & 0.981 $\pm$ 0.003 & \textbf{0.982} $\pm$ 0.000 \\
        osimertinib\_mpo & \textbf{0.876} $\pm$ 0.003 & \textbf{0.876} $\pm$ 0.008 & 0.873 $\pm$ 0.003 & \textbf{0.876} $\pm$ 0.008 & \textbf{0.876} $\pm$ 0.008 \\
        perindopril\_mpo & 0.697 $\pm$ 0.014 & 0.703 $\pm$ 0.006 & 0.703 $\pm$ 0.000 & 0.703 $\pm$ 0.009 & \textbf{0.718} $\pm$ 0.012 \\
        qed & 0.927 $\pm$ 0.000 & \textbf{0.942} $\pm$ 0.000 & 0.941 $\pm$ 0.000 & \textbf{0.942} $\pm$ 0.000 & \textbf{0.942} $\pm$ 0.000 \\
        ranolazine\_mpo & 0.809 $\pm$ 0.009 & 0.818 $\pm$ 0.016 & \textbf{0.824} $\pm$ 0.012 & 0.821 $\pm$ 0.011 & 0.821 $\pm$ 0.011 \\
        scaffold\_hop & 0.617 $\pm$ 0.002 & 0.621 $\pm$ 0.012 & 0.626 $\pm$ 0.005 & \textbf{0.628} $\pm$ 0.008 & \textbf{0.628} $\pm$ 0.008 \\
        sitagliptin\_mpo & 0.573 $\pm$ 0.006 & 0.560 $\pm$ 0.037 & 0.566 $\pm$ 0.010 & 0.573 $\pm$ 0.050 & \textbf{0.584} $\pm$ 0.034 \\
        thiothixene\_rediscovery & 0.650 $\pm$ 0.073 & 0.686 $\pm$ 0.121 & 0.677 $\pm$ 0.122 & 0.687 $\pm$ 0.125 & \textbf{0.692} $\pm$ 0.123 \\
        troglitazone\_rediscovery & 0.801 $\pm$ 0.062 & 0.853 $\pm$ 0.035 & 0.832 $\pm$ 0.052 & \textbf{0.867} $\pm$ 0.022 & \textbf{0.867} $\pm$ 0.022 \\
        valsartan\_smarts & 0.739 $\pm$ 0.043 & 0.797 $\pm$ 0.036 & 0.764 $\pm$ 0.038 & 0.797 $\pm$ 0.033 & \textbf{0.822} $\pm$ 0.042 \\
        zaleplon\_mpo & 0.406 $\pm$ 0.002 & 0.569 $\pm$ 0.014 & \textbf{0.586} $\pm$ 0.011 & 0.569 $\pm$ 0.005 & 0.584 $\pm$ 0.011 \\
    \hline
        \rule{0pt}{10pt}Sum & 17.641 & 18.091 & 18.074 & 18.208 & \textbf{18.362} \\
    \Xhline{0.2ex}
    \end{tabular}}
    \label{tab:app_pmo_ablation}
\end{table*}

\begin{table*}[t!]
    \caption{\small \textbf{Ablation study on goal-directed lead optimization (kcal/mol)} with $\delta=0.4$. The results are the mean docking scores of the most optimized leads of 3 runs. Lower is better and the best results are highlighted in bold.}
    \centering
    \resizebox{0.95\textwidth}{!}{
    \begin{tabular}{cc|ccccc}
    \Xhline{0.2ex}
        \rule{0pt}{10pt}Target protein & Seed score & Attaching (A) & \makecell{A +\\Token remasking} & \makecell{A +\\GPT remasking} & \makecell{A +\\Frag. remasking (F)} & A + F + MCG \\
    \hline
        \rule{0pt}{10pt} & -7.3 & -8.2 & -8.7 & -10.4 & \textbf{-10.6} & \textbf{-10.6} \\
        parp1 & -7.8 & -8.3 & -8.2 & \textbf{-11.5} & -11.0 & -11.0 \\
        & -8.2 & - & -10.9 & -11.1 & -10.9 & \textbf{-11.3} \\
    \hline
        \rule{0pt}{10pt} & -6.4 & -7.2 & -7.8 & -8.1 & \textbf{-8.4} & \textbf{-8.4} \\
        fa7 & -6.7 & -8.3 & -8.0 & -8.1 & \textbf{-8.4} & \textbf{-8.4} \\
        & -8.5 & - & - & - & - & - \\
    \hline
        \rule{0pt}{10pt} & -4.5 & - & -12.8 & -12.1 & \textbf{-12.9} & \textbf{-12.9} \\
        5ht1b & -7.6 & -11.9 & -11.8 & -12.1 & \textbf{-12.3} & \textbf{-12.3} \\
        & -9.8 & - & -11.3 & -11.2 & \textbf{-11.8} & \textbf{-11.8} \\
    \hline
        \rule{0pt}{10pt} & -9.3 & -9.7 & -10.7 & \textbf{-10.8} & \textbf{-10.8} & \textbf{-10.8} \\
        braf & -9.4 & - & \textbf{-10.8} & -10.1 & -10.2 & \textbf{-10.8} \\
        & -9.8 & - & -10.5 & \textbf{-10.6} & \textbf{-10.6} & \textbf{-10.6} \\
    \hline
        \rule{0pt}{10pt} & -7.7 & -8.6 & -8.7 & \textbf{-10.2} & -10.0 & \textbf{-10.2} \\
        jak2 & -8.0 & -8.9 & -9.0 & \textbf{-10.1} & -9.9 & -10.0 \\
        & -8.6 & - & -9.2 & \textbf{-9.8} & -9.5 & \textbf{-9.8} \\
    \Xhline{0.2ex}
    \end{tabular}}
    \label{tab:app_lead_ablation}
\end{table*}

\clearpage

\end{document}